%% file: neurips_2025.tex
\documentclass{article}

 \usepackage[dandb, final]{neurips_2025}

\usepackage[utf8]{inputenc} 
\usepackage[T1]{fontenc}    
\usepackage{hyperref}       
\usepackage{url}            
\usepackage{booktabs}       
\usepackage{amsfonts}       
\usepackage{nicefrac}       
\usepackage{microtype}      
\usepackage{xcolor}         
\usepackage{amsmath}
\usepackage{multirow}
\usepackage{booktabs}
\usepackage{graphicx}
\usepackage{wrapfig}
\usepackage{bm}
\usepackage{titletoc}

\newcommand{\eg}{\textit{e.g.}}
\newcommand{\ie}{\textit{i.e.}}

\title{CAPability: A Comprehensive Visual Caption \\ Benchmark for Evaluating \\ Both Correctness and Thoroughness}

\author{
Zhihang Liu\textsuperscript{1}\thanks{Interns at Tongyi Lab, Alibaba Group},
Chen-Wei Xie\textsuperscript{2},
Bin Wen\textsuperscript{2},
Feiwu Yu\textsuperscript{2},
Jixuan Chen\textsuperscript{2},
Pandeng Li\textsuperscript{1,2}\thanks{Corresponding author}, \\
\textbf{Boqiang Zhang\textsuperscript{1},
Nianzu Yang\textsuperscript{3},
Yinglu Li\textsuperscript{1},
Zuan Gao\textsuperscript{1},
Yun Zheng\textsuperscript{2},
Hongtao Xie\textsuperscript{1}}
\\[0.6ex]
\textsuperscript{1~}University of Science and Technology of China\quad
\textsuperscript{2~}Tongyi Lab, Alibaba Group\quad
\textsuperscript{3~}SJTU\quad \\ [1ex]
Project Page: \href{https://capability-bench.github.io}{https://capability-bench.github.io} \\
}

\begin{document}

\maketitle

\input{paper.tex}

{
    \small
    \bibliographystyle{unsrt}
    \bibliography{main}
}

\input{checklist.tex}

\input{suppl.tex}


\end{document}

%% file: paper.tex
\maketitle

\begin{figure}[!h]
    \vspace{-20pt}
    \includegraphics[width=1.0\linewidth]{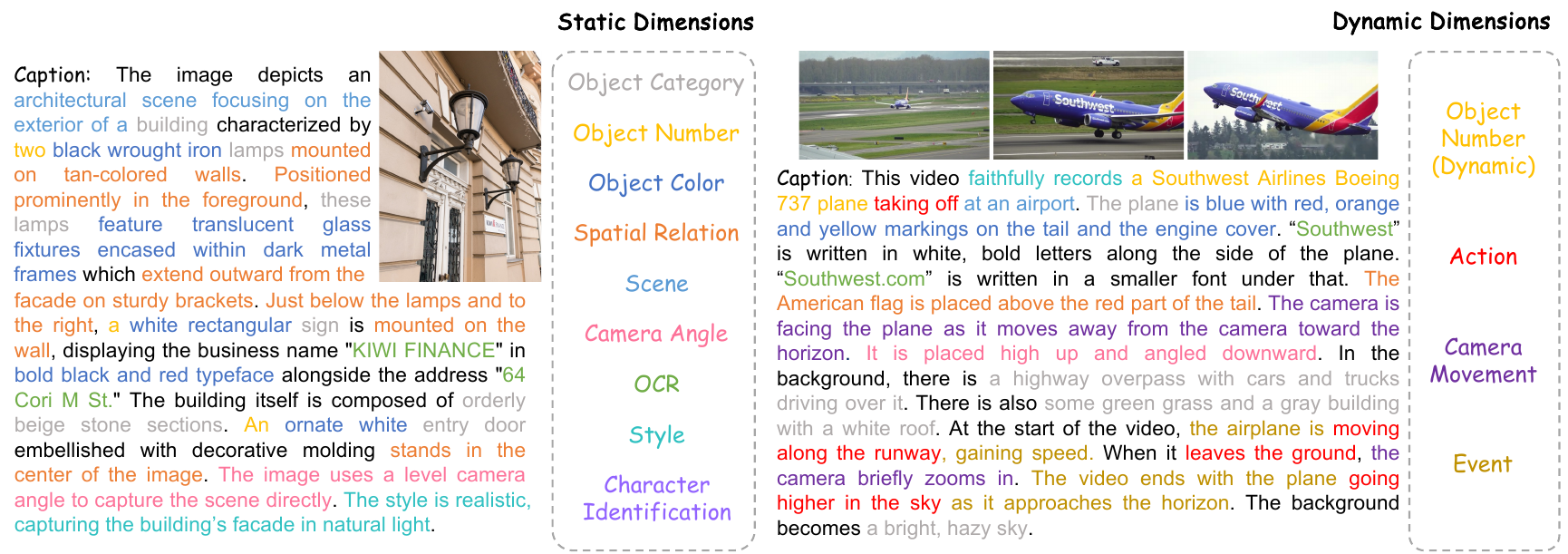}
    \caption{
        An example of image caption (left) and video caption (right) task. By analyzing the components of captions, we conclude 12 dimensions (9 static dimensions and 4 dynamic dimensions with object number shares on both static and dynamic), which all contribute to a detailed and comprehensive caption. The static dimensions are shared in both images and videos. Video data has additional dynamic dimensions that need to be judged with temporal relations.
    }
    \label{fig:teaser}
    \vspace{3pt}
\end{figure}

\begin{abstract}
Visual captioning benchmarks have become outdated with the emergence of modern multimodal large language models (MLLMs), as the brief ground-truth sentences and traditional metrics fail to assess detailed captions effectively. While recent benchmarks attempt to address this by focusing on keyword extraction or object-centric evaluation, they remain limited to vague-view or object-view analyses and incomplete visual element coverage. In this paper, we introduce CAPability, a comprehensive multi-view benchmark for evaluating visual captioning across 12 dimensions spanning six critical views. We curate nearly 11K human-annotated images and videos with visual element annotations to evaluate the generated captions. CAPability stably assesses both the correctness and thoroughness of captions with \textit{precision} and \textit{hit} metrics. By converting annotations to QA pairs, we further introduce a heuristic metric, \textit{know but cannot tell} ($K\bar{T}$), indicating a significant performance gap between QA and caption capabilities. Our work provides a holistic analysis of MLLMs' captioning abilities, as we identify their strengths and weaknesses across various dimensions, guiding future research to enhance specific aspects of their capabilities.
\end{abstract}

\section{Introduction}
\label{sec:intro}
Visual captioning, which translates visual content into textual descriptions, is a fundamental task for both image and video understanding, affecting various downstream tasks~\cite{jiang2023efficient, liu2024towards, jiang2025videopure}. The caption capability directly reflects the modality alignment ability~\cite{fang2015captions, xu2015show}, forms a significant basis for image and video generation~\cite{peebles2023scalable, wang2023modelscope, wei2024dreamvideo, wei2025dreamrelation}, and provides the precondition of synthesizing a large-scale multi-modal understanding and reasoning dataset~\cite{llavavideo, arnab2025temporal}. To assess the capabilities of this task, researchers established several visual caption benchmarks in earlier years~\cite{mscoco, nocaps, msrvtt, vatex}.

With the rapid development of recent MLLMs~\cite{llava, minigpt4, gpt4v, llavaov, gemini1.5, internvl2.5, nvila, videollama3, qwen2.5vl}, these traditional benchmarks have rapidly become outdated. This can be attributed to two main reasons: 1) The ground truths of traditional benchmarks often contain short sentences, missing many details. In contrast, recent MLLMs can produce much more detailed and fine-grained captions. 2) Traditional benchmarks use N-gram-based metrics (\eg, BLEU~\cite{papinesi2002bleu}, CIDER~\cite{vedantam2015cider}) to directly compare the similarity between sentences, making evaluations unreliable due to their high sensitivity to sentence style.

\begin{figure}[!t]
\centering
\includegraphics[width=0.9\textwidth]{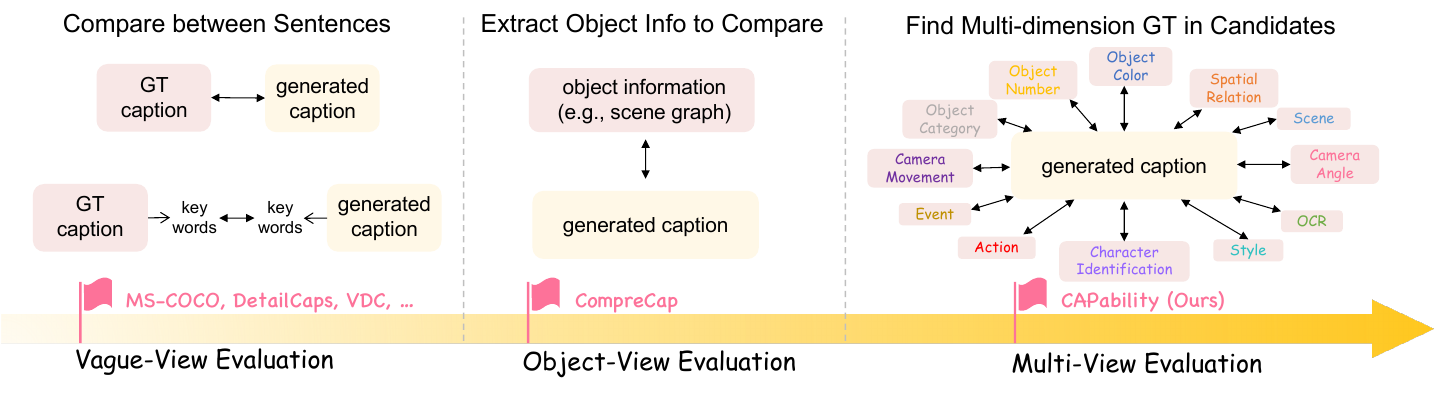}
\vspace{-8pt}
\caption{The development of visual caption benchmarks. Many works compare the ground-truth with generated sentences, which is vague. CompreCap~\cite{comprecap} uses a scene graph to evaluate only object-related information. Our CAPability considers multiple views with a comprehensive evaluation.}
\label{fig:motivation}
\vspace{-8pt}
\end{figure}

\begin{wraptable}{r}{0.45\textwidth}
\centering
\caption{Our designed views and more detailed dimensions. We can treat a caption from the listed six views, and then split them into several dimensions.}
\label{tab:dimensions}
\resizebox{\linewidth}{!}{%
\begin{tabular}{@{}ll@{}}
\toprule
\textbf{Views} & \textbf{Dimensions} \\ \midrule
Object-Related & \begin{tabular}[c]{@{}l@{}}Object Category, Object Color, \\ (Dynamic) Object Number, Spatial Relation\end{tabular} \\
Global-Related & Scene, Style \\
Text-Related & OCR \\
Camera-Related & Camera Angle, Camera Movement \\
Temporal-Related & Action, Event \\
Knowledge-Related & Character Identification \\ \bottomrule
\end{tabular}%
}
\vspace{-6pt}
\end{wraptable}
Recently, new visual caption benchmarks have been introduced to update the outdated ones. As illustrated in Fig.~\ref{fig:motivation}, Dream-1K~\cite{dream1k}, DetailCaps~\cite{detailcaps} and VDC~\cite{auroracap} extract keywords from both generated and reference captions, (\eg, objects in DetailCaps, events in Dream-1K, objects, background, and camera information in VDC), and then compare the extracted information to score the caption, as opposed to earlier methods that directly compared sentences. We name all these methods vague-view evaluation as their evaluations still depend on the level of detail and accuracy of the ground-truth caption, which can suffer from human bias and cumulative errors from repeated extraction and comparison by LLMs. 
CompreCap~\cite{comprecap} extracts object-related annotations from images (\eg, scene graph) without ground-truth captions, thereby focusing on evaluating the object description capabilities of modern image MLLMs. We refer to this as object-view evaluation, as it drops entire sentences as ground truth, and evaluates captions based on object representation. Compared to traditional benchmarks, all these newly introduced approaches aim to provide more precise ground truths and evaluation methods, enhancing the reliability and interpretability of benchmarking.

However, the evaluation of these benchmarks remains incomplete as they focus on a single aspect of captions with limited visual elements, inadequately covering the full caption scope. 
For instance, they often overlook aspects like scene, text, and style. 
We argue that a multi-view evaluation for visual caption is essential.
In this paper, we introduce a new comprehensive visual caption benchmark, CAPability. Our approach uses complete visual elements rather than caption sentences as annotations to evaluate both correctness and thoroughness for each dimension. 
The selection of dimensions is motivated by existing visual generation benchmarks~\cite{geneval, vbench, t2vcompbench}, which is the inverse task of visual captioning. They usually treat and evaluate the generated visual content from different aspects (\eg, objects, scene, style, camera, motion control). Similarly, we design 6 views and 12 dimensions for CAPability by analyzing several visual captions, as illustrated in Tab.~\ref{tab:dimensions} and Fig.~\ref{fig:teaser}.
We believe these components contribute to a complete caption, as lacking any of them may align the caption with different visual content. There are 9 static and 4 dynamic dimensions, with \textit{object number} encompassing both static and dynamic aspects. 
Static dimensions apply to both images and videos, while dynamic ones are exclusive to video. In our CAPability, we collect video data only for dynamic dimensions and image data for static dimensions for simplicity. We then manually annotate 11K images and videos for CAPability, providing sufficient samples.

\begin{figure}[!t]
\centering
\includegraphics[width=0.85\textwidth]{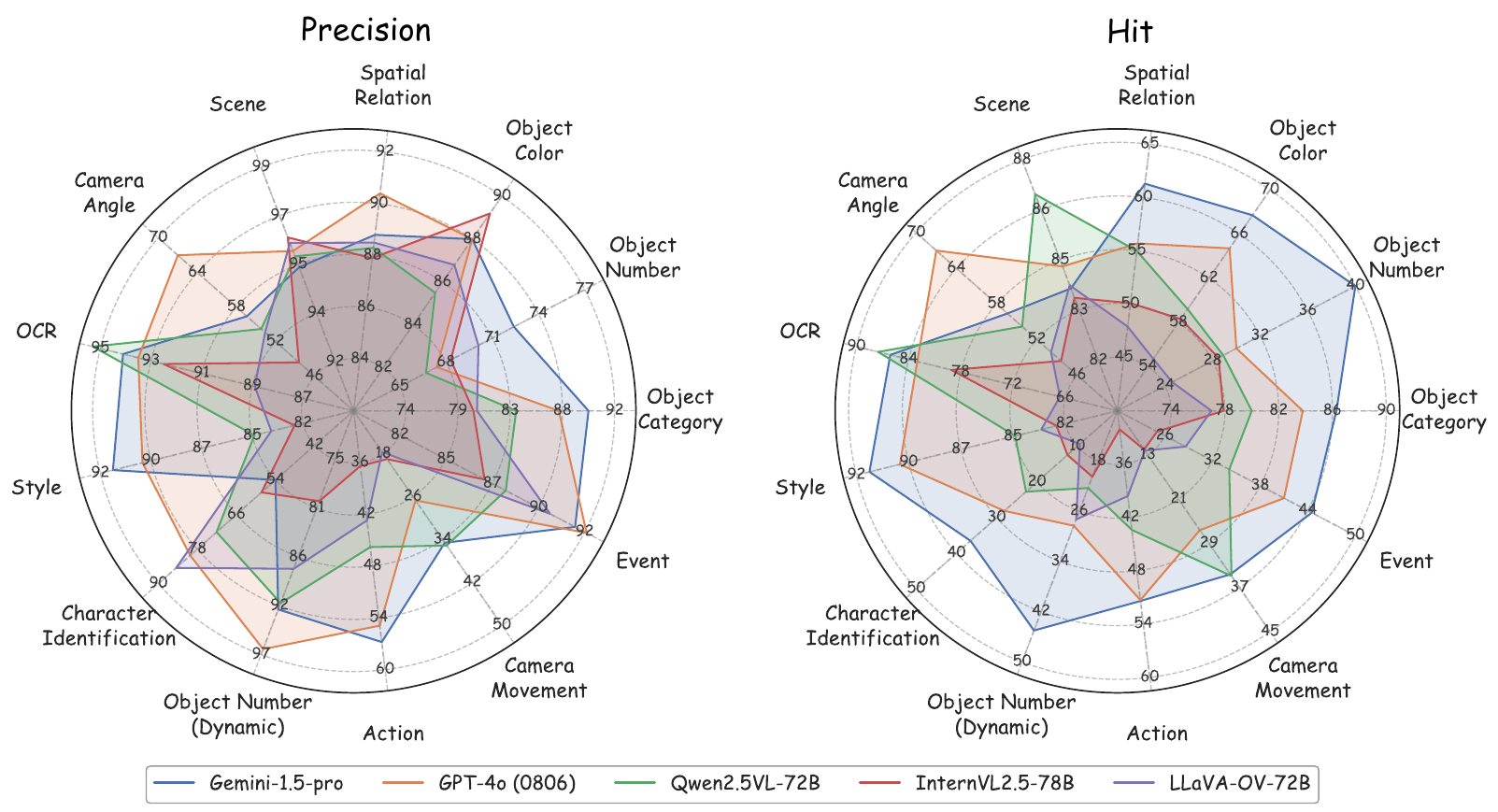}
\vspace{-5pt}
\caption{\textit{Precision} and \textit{hit} comparison of SOTA MLLMs on our CAPability. Models perform more variably on \textit{hit} metric, which evaluates the thoroughness. GPT-4o performs the best on \textit{precision}, and Gemini-1.5-pro~\cite{gemini1.5} performs the best on \textit{hit}.}
\label{fig:radar}
\vspace{-8pt}
\end{figure}

\begin{wraptable}{r}{0.4\textwidth}
\setlength\tabcolsep{3pt}
\centering
\caption{Comparison of our CAPability and other visual caption benchmarks in different aspects. We are the most comprehensive with image and video data, multi-view annotations, and new thoroughness evaluation methods proposed.}
\label{tab:benchmark_comparison}
\resizebox{\linewidth}{!}{%
\begin{tabular}{@{}l|cc|c|cc@{}}
\toprule
\multirow{2}{*}{\textbf{Benchmark}} & \multicolumn{2}{c|}{\textbf{Data Type}} & \multirow{2}{*}{\textbf{\begin{tabular}[c]{@{}c@{}}Anno-\\ tations\end{tabular}}} & \multicolumn{2}{c}{\textbf{Evaluation}} \\ \cline{2-3} \cline{5-6}
 & \textbf{Image} & \textbf{Video} &  & \textbf{Thoroughness} & $\bm{K\bar{T}}$ \\ \midrule
MS-COCO\cite{mscoco} & \checkmark & - & Sentences & - & - \\
MSRVTT~\cite{msrvtt} & - & \checkmark & Sentences & - & - \\
Dream-1K~\cite{dream1k} & - & \checkmark & Sentences & single dim & - \\
VDC~\cite{auroracap} & - & \checkmark & Sentences & - & - \\
DetailCaps~\cite{detailcaps} & \checkmark & - & Sentences & - & - \\
CompreCap~\cite{comprecap} & \checkmark & - & Object Info & single dim & - \\ \midrule
CAPability (Ours) & \checkmark & \checkmark & \begin{tabular}[c]{@{}c@{}}Multi-view\\ Elements\end{tabular} & \checkmark & \checkmark \\ \bottomrule
\end{tabular}%
}
\end{wraptable}

In addition to multi-view annotation, we also focus on improving the evaluation quality of captions. While most methods assess only the correctness, we argue that considering both correctness and thoroughness of visual elements provides a more comprehensive evaluation for visual captioning.
Therefore, we conduct comprehensive experiments using both \textit{precision} and \textit{hit} as our main metric, supplemented by another heuristic metric: the \textit{know but cannot tell} ($K\bar{T}$). While \textit{precision} only evaluates the accuracy of captions, \textit{hit} considers both accuracy and coverage of captions compared with ground truth, and $K\bar{T}$ indicates the ratio when a model can answer related questions correctly but fails to convey the same information in the caption automatically. These metrics provide a robust framework for evaluating both correctness and thoroughness in MLLMs' detailed captions.
To our knowledge, we are the first to heuristically highlight the gap in correctness and thoroughness capabilities of MLLMs across multiple views, providing deeper insight into specific capabilities or limitations of a model, thereby offering actionable guidance for further research and development, rather than just an overall score.
Representative results are shown in Fig.~\ref{fig:radar}, leading to the following conclusions: 
1) Models perform more variably on \textit{hit}, which evaluates the thoroughness and may be ignored by previous research.
2) GPT-4o seems the best on \textit{precision}. For \textit{hit}, Gemini-1.5-pro~\cite{gemini1.5} has a leading advantage in many dimensions, followed by GPT-4o~\cite{gpt4o}.
3) Gemini-1.5-pro demonstrates strong object-counting abilities, while GPT-4o excels in identifying camera angles.
4) All models still struggle with dimensions like object numbers, camera angle, camera movement, character identification, and action. We hope our findings guide researchers to focus on improving these abilities in caption tasks.
Our main contributions are listed as follows:
\begin{itemize}
\item \noindent We introduce a new comprehensive visual caption benchmark, CAPability, featuring 6 views and 12 dimensions. By collecting and human-annotating nearly 11K images and videos, CAPability provides a novel and comprehensive methodology for caption benchmarking.
\item \noindent We emphasize that captions should be evaluated for both correctness and thoroughness. Accordingly, we report \textit{precision} and \textit{hit} to combine correctness and thoroughness.
\item \noindent We transform our annotations into a QA format to evaluate QA accuracy. Based on this approach, we assess an additional capability via the $K\bar{T}$ metric, which indicates the performance gap between QA and the captioning task.
\end{itemize}

\section{Related Work}
\noindent\textbf{Multi-modal large language models.}
Based on the significant development of Large Language Models (LLMs) among various linguistic tasks~\cite{gpt3, vicuna, qwen2.5, llama3}, many works try to extend the powerful capabilities into multi-modal understanding~\cite{liu2025hybrid, bao2024cores, bao2025dynimg}. By integrating image content into LLMs, Multi-modal Large Language Models (MLLMs) also gain huge achievements~\cite{llava, llava1.5, qwenvl, minigpt4, llavanext, internvl, gpt4v}. Based on the pre-trained weights from image models, recent MLLMs also expand video understanding capabilities~\cite{llavanextvideo, llavaov, llavavideo, qwen2vl, internvl2.5, nvila, videollama3, qwen2.5vl}. With rapid development, MLLMs are powerful enough to describe both the image and video content in detail, which makes the traditional benchmarks with short captions outdated. More and more methods even try to produce re-captioned detailed descriptions by more powerful models rather than existing human-annotated short captions to train their model~\cite{llavaov, videollama3}. Therefore, it is urgent to propose a new visual caption benchmark that adapts to modern MLLMs.

\noindent\textbf{Visual caption benchmarks.}
Visual Caption is a fundamental task in computer vision. Early visual caption benchmarks, such as MS-COCO~\cite{mscoco}, NoCaps~\cite{nocaps}, MSR-VTT~\cite{msrvtt}, and VATEX~\cite{vatex}, usually contain a short sentence with limited visual information as the ground truth. They also use metrics like BLEU~\cite{papinesi2002bleu}, CIDER~\cite{vedantam2015cider}, and METEOR~\cite{banerjee2005meteor} to calculate the matching score directly between two sentences, which is easily affected by the sentence style. Recently, from the annotation aspect, DetailCaps~\cite{detailcaps} extracts object-related information from the ground-truth caption, Dream-1K~\cite{dream1k} splits the ground truth and candidates sentences into events. VDC~\cite{auroracap} also extracts the object, background, and camera information from the video captions by question templates. However, they still rely on the ground-truth caption with human-bias, and require existing LLMs to extract and compare multiple times, thus increasing the cumulative error. CompreCap~\cite{comprecap} explores directly annotating the object-related information in image captions, making the benchmarking more interpretable. On the contrary, we are the first time to propose a comprehensive visual caption benchmark covering both image and video data with 6 views and 12 dimensions. For evaluation, most methods only focus on correctness. Dream-1K~\cite{dream1k} and CompreCap~\cite{comprecap} begin to focus on thoroughness and calculate the recall of events or the object coverage in the segmentation map. However, they still remain incomplete as they only evaluate one dimension and limited metrics. We design metrics about both correctness and thoroughness, which may be ignored by previous work. We summarize the comparison with other visual caption benchmarks in Tab.~\ref{tab:benchmark_comparison}, and we are the most holistic on all listed aspects.

\begin{figure}[!t]
\centering
\includegraphics[width=0.88\textwidth]{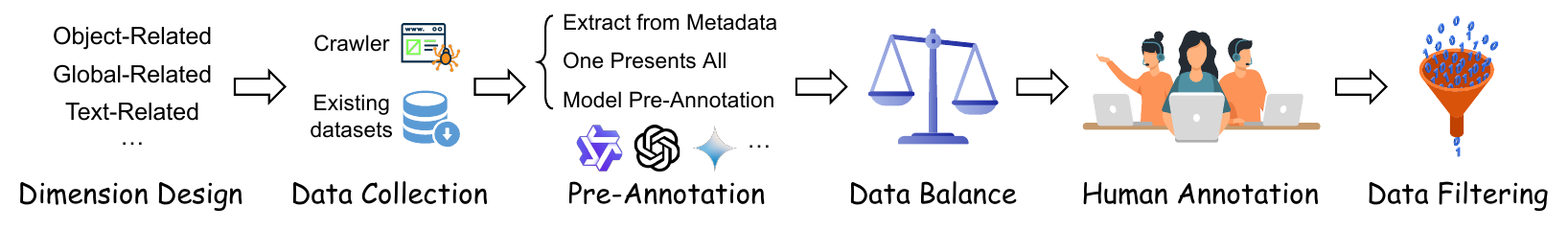}
\vspace{-4pt}
\caption{The pipeline of our data annotation for each dimension.}
\label{fig:anno_pipeline}
\vspace{-8pt}
\end{figure}

\section{CAPability}
\subsection{Multiple Dimension Data Annotation}
\label{sec:capability_annotation}
The pipeline of our whole collection and annotation is shown in Fig.~\ref{fig:anno_pipeline}. We first design 6 views and split 12 dimensions, then collect nearly 1,000 images and videos for each dimension with low overlap among dimensions. For the collected data, we conduct pre-annotations by SOTA MLLMs and the following data balancing before the human annotation. After the human annotation of each dimension, we filter bad cases during the annotation and finally complete the data of CAPability.

\begin{figure}[!t]
\centering
\includegraphics[width=0.95\textwidth]{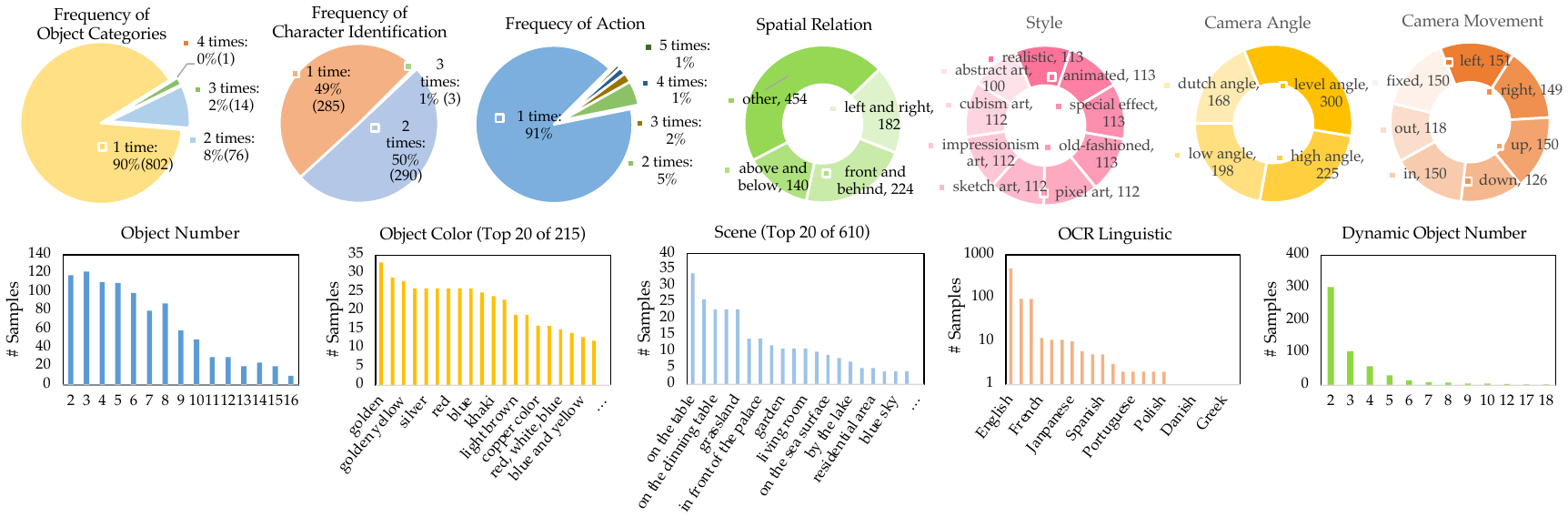}
\vspace{-8pt}
\caption{The annotation distribution of each dimension. We statistic different dimensions with different types. We count the frequency in object categories, character identification, and action as most of the descriptions only appear one time. For spatial relation, we summarize 4 categories and count their numbers. For style, camera angle, and camera movement, we count the samples of each category. For others, we plot bar charts to count and show the most frequent samples.}
\label{fig:gt_distribution}
\end{figure}

\noindent\textbf{Dimension design.}
Motivated by visual generation benchmarks~\cite{geneval, vbench, t2vcompbench}, we conclude 6 views and split them into 12 dimensions based on the analysis of caption cases, as shown in Tab.~\ref{tab:dimensions}. As shown in Fig.~\ref{fig:teaser}, we design 9 static dimensions for both video and image, and 4 dynamic dimensions for video, covering most of what makes up a visual caption. We classify dimensions as dynamic or static based on the following principle: descriptions obtainable from a single frame are static, those requiring the entire video are dynamic, which are more related to temporal information. For the object number dimension, the number can be counted statically in an image, and also dynamically in a video, which is more challenging~\cite{vsibench}.
We also design the annotation type of each dimension as two types: open-ended, and specific categories. Specifically, we define 9 categories for style, 4 categories for camera angle, and 7 categories for camera movement. The specific categories of each dimension can be found in Fig.~\ref{fig:gt_distribution}. See Appendix~\ref{sec:supp_dimension} for details of dimension design.

\begin{figure}[!t]
\centering
\includegraphics[width=0.95\textwidth]{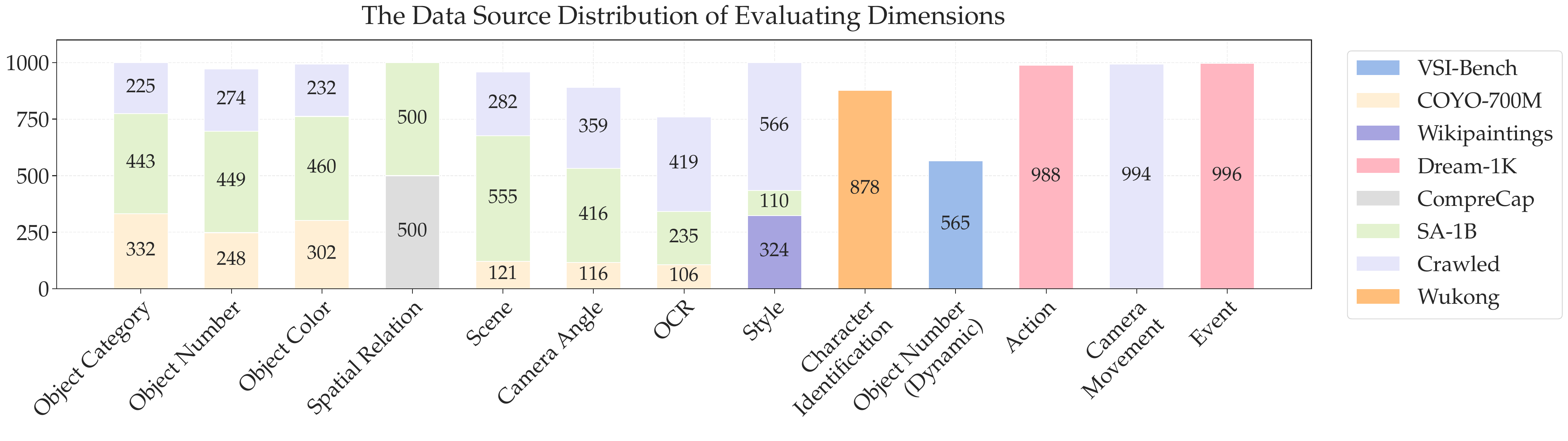}
\vspace{-8pt}
\caption{The data source count and distribution of each dimension. We collect nearly 1,000 images/videos for each dimension, crawl parts of data by ourselves, and sample some data from existing datasets to ensure diversity.}
\label{fig:data_source}
\end{figure}

\noindent\textbf{Data collection.}
For convenience and problem simplification, we only collect image data for static dimensions and video data for dynamic dimensions.
This is based on the common sense that the video understanding capabilities for MLLMs are usually built upon sufficient image understanding capabilities~\cite{llavanextvideo, llavavideo, videollama3, internvl2.5}. 
Since an image or a video cannot cover all these dimensions of information, we directly collect data for each dimension independently and evaluate each dimension separately.
For static dimensions, we mainly collect images from SA-1B~\cite{sam}, COYO-700M~\cite{coyo700m}, Wukong~\cite{wukong100m}, and Wikipaintings~\cite{wikipaintings}, and we also crawl a considerable amount of data from multiple public datasets and websites with CC0 license by ourselves. We also borrow parts of the image data and annotations from CompreCap~\cite{comprecap} for the spatial relation dimension. For dynamic dimensions, we crawl and cut videos for camera movement dimension, borrow videos from Dream-1K~\cite{dream1k} for action and event dimensions, and borrow videos from VSI-Bench~\cite{vsibench} for the dynamic object number dimension. Fig.~\ref{fig:data_source} shows our data sources for each dimension and their proportion, and Tab.~\ref{tab:data_overlap} shows our data overlap among dimensions.

\noindent\textbf{Pre-Annotation.}
The annotations may not be unique for different dimensions. For global-related, camera-related, and knowledge-related views, the annotations tend to be unique as an image only belongs to one kind of scene, style, \textit{etc}. We directly pre-annotate them by extracting the metadata (\eg, style for images in Wikipaintings~\cite{wikipaintings}), or ask SOTA MLLMs to get a preliminary answer. For object-related, text-related, and temporal-related views, there could be multiple objects, texts, or actions in an image or video.
However, it is extremely hard to annotate all objects or actions within an image or a video, as the categories of objects can be divided by almost infinite granularity~\cite{tang2023visual, wang2023all, wang2024all}. Therefore, we do not pursue the most comprehensive annotation possible for each single sample, but randomly select only one object from the visual content, and the same for other dimensions, and reflect the accuracy and thoroughness through the evaluation of a large number of samples. We name this strategy as \textit{One Represents All}. According to the law of large numbers, the distribution of randomly selection can approximate the expectation of covering different granularities of the entire visual content with a large amount of samples, thus ensuring the unbiased nature of the benchmark. Therefore, the key of this annotation strategy is to keep the selection as random as possible. To avoid humans' bias on selecting, we ask the three SOTA MLLMs, \ie, GPT-4o~\cite{gpt4o}, Gemini-1.5-pro~\cite{gemini1.5}, and Qwen-VL-Max~\cite{qwen2vl} to list all objects and actions at the granularity they deem appropriate in an image or video, ask PaddleOCR~\cite{ppocrv3} to list all texts in an image. We finally use Qwen2.5-Max~\cite{qwen2.5} to merge the results together and randomly select one from the merged list to obtain the pre-annotated results. For further object-related dimensions, \eg, object number, object color, and spatial relation, the object selection follows this strategy, then pre-annotate these attributes by MLLMs. 

\noindent\textbf{Data balance.} Based on the pre-annotation results for each dimension, we conduct data balance strategy to control the difficulty and diversity. For dimensions with specific categories, we try to make the number of each category similar. For dimensions of open-ended descriptions, we count the frequency of descriptions, suppress the long-tail distribution, keep low-frequency words, ensuring the variety. See Appendix~\ref{sec:supp_data_balance} for examples. The final annotation distribution is shown as Fig.~\ref{fig:gt_distribution}.

\noindent\textbf{Human annotation.}
For different dimensions, we design different tasks for human annotators. For example, human annotators are asked to judge only right or wrong for object categories and actions rather than changing the annotated descriptions since we need to keep randomness. For dimensions with specific categories, we ask annotators to check the pre-annotated option and select the correct option. As for other dimensions with open-ended descriptions, we ask annotators to check the pre-annotations one by one and modify them based on pre-defined rules if there are any mistakes. We also conduct human-validation of all annotations to ensure the accuracy of annotations is above 97\%.

\noindent\textbf{Data filtering.}
We finally conduct data filtering to drop harmful visual content and re-balance the data since many of them are modified manually. The final distribution of each dimension is shown in Fig.~\ref{fig:gt_distribution}. Some benchmark examples are shown in Appendix~\ref{sec:supp_example}.

\subsection{Multiple Dimension Evaluation}
\noindent\textbf{Caption evaluation.}
As we collect and annotate the data of each dimension separately, we also evaluate each dimension independently.
Different from matching the similarity between the caption and ground-truth sentences, our annotation drops the caption sentence, and we use GPT-4 Turbo~\cite{gpt4} to take interpretable scores for each dimension. We use a similar prompt template for dimensions with specific categories (\ie, style, camera angle, and camera movement), and use another similar prompt template for other open-set dimensions. See Appendix~\ref{sec:supp_prompt} for the details of the prompts. Therefore, we can judge the caption into the following three situations: 
1) \textbf{MIS}, which means the caption does not mention the corresponding content about the dimension.
2) \textbf{COR}, which means the caption mentions the corresponding content about the dimension, and describes it correctly compared with the annotations.
3) \textbf{INC}, which means the caption mentions the corresponding content about the dimension, but gives a wrong description compared with the annotations.
As all data can be judged into these three situations, we can then calculate two metrics: 
1) \textit{\textbf{Precision}}, which represents the accuracy on all samples that the model has mentioned, and thus only considers the correctness. 2) \textit{\textbf{Hit}}, which represents the accuracy on all annotated samples, no matter whether the dimension is described or missed in the caption, and thus also considers thoroughness. Our \textit{Hit} is similar to recall, see Appendix~\ref{sec:supp_metrics} for more discussions.
Given the set of all samples as $S(\text{ALL})$, positive samples as $S(\text{COR})$, negative samples as $S(\text{INC})$, missed samples as $S(\text{MIS})$, the metrics can be calculated by:
\begin{align}
    \label{eq:precision}
    \text{Precision} &= \frac{|S(\text{COR})|}{|S(\text{COR}) \cup S(\text{INC})|}, \\
    \label{eq:hit}
    \text{Hit} &= \frac{|S(\text{COR})|}{|S(\text{ALL})|}.
\end{align}

\begin{table}[!t]
\setlength\tabcolsep{4pt}
\centering
\caption{The \textit{precision} and \textit{hit} of closed-source models and 72B open-source models on all dimensions. The \textit{precision} represents the accuracy of what the models have described. The \textit{hit} shows how many visual elements in the image can be described correctly. For video inputs, we send the whole video for Gemini, and uniformly sample 50 frames for GPT due to the API limitation of image number.}
\label{tab:72b_f1score}
\resizebox{0.9\textwidth}{!}{%
\begin{tabular}{@{}llcccccccccccccc@{}}
\toprule
 & \textbf{Methods} & \textbf{\begin{tabular}[c]{@{}c@{}}Obj.\\ Cate.\end{tabular}} & \textbf{\begin{tabular}[c]{@{}c@{}}Obj.\\ Num.\end{tabular}} & \textbf{\begin{tabular}[c]{@{}c@{}}Obj.\\ Color\end{tabular}} & \textbf{\begin{tabular}[c]{@{}c@{}}Spa.\\ Rel.\end{tabular}} & \textbf{Scene} & \textbf{\begin{tabular}[c]{@{}c@{}}Cam.\\ Ang.\end{tabular}} & \textbf{OCR} & \textbf{Style} & \textbf{\begin{tabular}[c]{@{}c@{}}Cha.\\ Iden.\end{tabular}} & \makebox[0.048\textwidth][c]{\textbf{\small{\begin{tabular}[c]{@{}c@{}}(D) Obj.\\ Num.\end{tabular}}}} & \textbf{Act.} & \textbf{\begin{tabular}[c]{@{}c@{}}Cam.\\ Mov.\end{tabular}} & \textbf{Event} & \textbf{Avg.} \\ \midrule
\multirow{7}{*}{\rotatebox{90}{\textit{Precision}}} & LLaVA-OV-72B & 80.4 & 70.1 & 86.8 & 88.5 & 96.2 & 53.6 & 88.9 & 83.9 & \textbf{84.5} & 87.5 & 42.7 & 17.7 & 90.2 & 74.7 \\
 & Qwen2VL-72B & 82.0 & 70.0 & 89.2 & 88.6 & 95.2 & 52.4 & \textbf{95.9} & 82.9 & 83.3 & 83.6 & 44.6 & 33.3 & 89.9 & 76.2 \\
 & InternVL2.5-78B & 80.1 & 68.3 & 89.2 & 87.9 & \textbf{96.4} & 48.4 & 92.5 & 82.8 & 58.3 & 80.0 & 36.4 & 19.0 & 86.8 & 71.2 \\
 & Qwen2.5VL-72B & 83.7 & 66.7 & 85.5 & 88.3 & 95.7 & 54.2 & 95.3 & 84.7 & 72.1 & 91.3 & 45.8 & 35.1 & 87.9 & 75.9 \\
 & GPT-4o (0806) & 87.3 & 67.4 & 88.0 & \textbf{90.4} & 95.9 & \textbf{67.0} & 93.5 & 90.0 & 80.2 & \textbf{96.4} & 54.9 & 26.7 & \textbf{92.1} & 79.2 \\
 & Gemini-1.5-pro & \textbf{89.8} & 72.4 & 88.0 & 88.8 & 95.3 & 56.4 & 94.1 & \textbf{91.4} & 54.0 & 92.0 & \textbf{56.8} & 34.6 & 91.5 & 77.3 \\
 & Gemini-2.0-flash & 85.9 & \textbf{78.6} & \textbf{90.4} & 89.0 & 96.1 & 57.4 & 95.3 & 86.9 & 82.0 & 94.2 & 50.6 & \textbf{35.4} & 89.0 & \textbf{79.3} \\ \midrule
\multirow{7}{*}{\rotatebox{90}{\textit{Hit}}} & LLaVA-OV-72B & 77.0 & 24.6 & 54.7 & 47.9 & 84.0 & 49.9 & 66.6 & 83.5 & 9.3 & 27.3 & 39.6 & 12.2 & 28.6 & 46.6 \\
 & Qwen2VL-72B & 79.9 & 25.1 & 57.3 & 50.4 & 85.1 & 52.1 & 79.6 & 82.6 & 5.7 & 18.1 & 41.7 & 25.7 & 31.2 & 48.8 \\
 & InternVL2.5-78B & 77.9 & 28.5 & 58.3 & 50.1 & 83.6 & 48.4 & 79.1 & 82.8 & 12.4 & 20.5 & 32.1 & 12.0 & 25.0 & 47.0 \\
 & Qwen2.5VL-72B & 80.0 & 28.9 & 59.2 & 55.0 & \textbf{86.9} & 54.2 & 87.5 & 84.7 & 22.7 & 22.3 & 43.4 & 34.9 & 34.1 & 53.4 \\
 & GPT-4o (0806) & 83.8 & 30.0 & 64.7 & 55.7 & 84.6 & \textbf{67.0} & 83.0 & 90.0 & 28.1 & 28.3 & 51.3 & 26.6 & 41.0 & 56.5 \\
 & Gemini-1.5-pro & \textbf{86.3} & \textbf{40.0} & \textbf{67.7} & \textbf{61.3} & 83.9 & 56.4 & 86.1 & \textbf{91.4} & 36.5 & \textbf{45.0} & \textbf{51.4} & 34.6 & \textbf{44.5} & \textbf{60.4} \\
 & Gemini-2.0-flash & 82.5 & 30.6 & 60.8 & 51.8 & 84.0 & 57.4 & \textbf{88.8} & 86.8 & \textbf{37.9} & 28.7 & 46.6 & \textbf{35.2} & 39.7 & 56.2 \\ \bottomrule
\end{tabular}%
}
\vspace{-8pt}
\end{table}

\noindent\textbf{Question-answer pairs evaluation.}
As we annotate each descriptive element for each dimension rather than caption sentence, we can also convert our annotations to question-answer (QA) pair format to evaluate the MLLMs' general ability out of the horizon of caption only. See Appendix~\ref{sec:supp_example} for examples of our CAPability-QA. Based on the QA accuracy, we introduce a new metric, \textit{know but cannot tell} ($K\bar{T}$), which evaluates the situation when a model knows the answer (\ie, can answer correctly when given it the related question), but cannot tell automatically in the caption without specific question as prompt. This evaluation is significant to the caption task of MLLMs, but is usually ignored by previous methods. Given the set of correct answers as $S_{qa}(\text{COR})$, $K\bar{T}$ can be calculated as:
\begin{equation}
    K\bar{T} = \frac{\left|S_{qa}(\text{COR}) \cap [S(\text{INC}) \cup S(\text{MIS})]\right|}{|S_{qa}(\text{COR})|}.
\end{equation}

\section{Experiments}

\subsection{Experimental Setups}
For comprehensively evaluating the state-of-the-art (SOTA) models, we both choose several popular open-source and closed-source MLLMs. For closed-source models, we evaluate GPT-4o (0806)~\cite{gpt4o}, Gemini-1.5-pro~\cite{gemini1.5}, and Gemini-2.0-flash~\cite{gemini2.0}.
For open-source models, we evaluate InternVL2.5~\cite{internvl2.5}, LLaVA-OneVision~\cite{llavaov}, NVILA~\cite{nvila}, VideoLLaMA3~\cite{videollama3}, Qwen2VL~\cite{qwen2vl} and Qwen2.5VL~\cite{qwen2.5vl} with their different LLM sizes. We use the same image prompt and video prompt to infer all models, see Appendix~\ref{sec:supp_prompt} for the inference prompts. We use GPT-4 Turbo (1106-preview)~\cite{gpt4} to take scores for all generated captions to complete our evaluation. See Appendix~\ref{sec:supp_impl_details} for more implementation details.

\subsection{Main Evaluation Results}

\noindent\textbf{Precision and hit of closed-source API and 72B models.}
We report the \textit{precision} and \textit{hit} of closed-source and 72B models in Tab.~\ref{tab:72b_f1score}. Gemini-2.0-flash and GPT-4o achieve the highest \textit{precision} (79.3\% and 79.2\%), which represents their captions are truthful and accurate. When it comes to the \textit{hit} metric, all results drop significantly as it is harder to cover as many visual elements as possible. Gemini-1.5-pro achieves the best with 3.9\% higher than second place, \ie, GPT-4o, which means it is better at identifying more elements correctly. Among 72B models, Qwen2.5VL achieves the best \textit{hit}, with notable \textit{precision}. 
We note that the high precision but moderate hit for models like GPT-4o suggests they tend to describe only what they are confident about and avoid parts they are less certain of (line 245). This behavior results in high accuracy for described elements, but at the cost of thoroughness and coverage.
When we focus on each dimension, it is worth noting that these models behave differently in different dimensions. Gemini-1.5-pro has a huge advantage in object counting in both image and video, especially in thoroughness (10\% better than the second place for image and 16.3\% better than the second place for video in \textit{hit} metric). GPT-4o excels at recognizing camera angle, as it is 9.6\% higher than the second place on both \textit{precision} and \textit{hit}. Qwen2.5-VL performs the best \textit{hit} in the open-source models as it performs well on scene and camera movement. Object category, scene, OCR, and style seem simple for these powerful models, as they all achieve well on the F1-score. When it comes to the dimensions of object number, object color, spatial relation, style, character identification, and events, all of them show relatively high \textit{precision} but low \textit{hit}, which means they can describe these elements well when they are confident about them, but might miss some instances and ignore the thoroughness. As for the action and two camera-related dimensions, all models perform unsatisfactorily on both \textit{precision} and \textit{hit}.
This phenomenon inspires researchers to focus more on these aspects of the model's capability.
We acknowledge that this finding highlights well-known open challenges for current MLLMs. For example, accurate object counting requires fine-grained discrimination; some recent methods~\cite{deepeyes, jiang2025vlm} leverage segmentation or external tools to assist reasoning. Camera-related tasks demand a strong visual reasoning ability to infer subtle contextual cues. Character identification relies heavily on high-quality training data to build related knowledge. Action recognition is inherently challenging due to the temporal modeling.

\begin{table}[!t]
\setlength\tabcolsep{4pt}
\centering
\caption{The \textit{precision} and \textit{hit} of 7B open-source models on all dimensions. We keep their default settings for each model.}
\label{tab:7b_f1score}
\resizebox{0.9\textwidth}{!}{%
\begin{tabular}{@{}llcccccccccccccc@{}}
\toprule
 & \textbf{Methods} & \textbf{\begin{tabular}[c]{@{}c@{}}Obj.\\ Cate.\end{tabular}} & \textbf{\begin{tabular}[c]{@{}c@{}}Obj.\\ Num.\end{tabular}} & \textbf{\begin{tabular}[c]{@{}c@{}}Obj.\\ Color\end{tabular}} & \textbf{\begin{tabular}[c]{@{}c@{}}Spa.\\ Rel.\end{tabular}} & \textbf{Scene} & \textbf{\begin{tabular}[c]{@{}c@{}}Cam.\\ Ang.\end{tabular}} & \textbf{OCR} & \textbf{Style} & \textbf{\begin{tabular}[c]{@{}c@{}}Cha.\\ Iden.\end{tabular}} & \makebox[0.048\textwidth][c]{\textbf{\small{\begin{tabular}[c]{@{}c@{}}(D) Obj.\\ Num.\end{tabular}}}} & \textbf{Act.} & \textbf{\begin{tabular}[c]{@{}c@{}}Cam.\\ Mov.\end{tabular}} & \textbf{Event} & \textbf{Avg.} \\ \midrule
\multirow{6}{*}{\rotatebox{90}{\textit{Precision}}} & LLaVA-OV-7B & 79.5 & 67.8 & 87.3 & 88.7 & 95.3 & 41.9 & 87.7 & \textbf{84.4} & \textbf{90.9} & \textbf{92.8} & 38.9 & 20.2 & 87.3 & 74.0 \\
 & Qwen2VL-7B & 80.3 & 68.3 & \textbf{88.7} & 89.0 & 95.4 & 39.9 & \textbf{95.4} & 77.1 & 83.3 & 73.5 & 42.5 & 24.2 & 86.7 & 72.6 \\
 & NVILA-8B & 80.6 & 68.5 & 84.2 & 88.4 & 95.4 & 44.5 & 92.8 & 79.0 & 90.8 & 47.2 & 32.4 & 14.7 & \textbf{92.1} & 70.0 \\
 & InternVL2.5-8B & 76.1 & 60.3 & 85.8 & 89.3 & 95.1 & 42.5 & 89.0 & 81.9 & 48.3 & 84.4 & 37.9 & 20.2 & 84.5 & 68.9 \\
 & VideoLLaMA3-7B & 81.0 & 66.8 & 85.5 & \textbf{90.6} & \textbf{97.0} & 43.2 & 90.0 & 80.9 & 84.1 & 78.9 & 43.0 & \textbf{30.2} & 88.3 & 73.8 \\
 & Qwen2.5VL-7B & \textbf{82.0} & \textbf{73.5} & 88.0 & 88.6 & 95.8 & \textbf{47.7} & 93.8 & 82.0 & 80.8 & 92.4 & \textbf{43.7} & 26.8 & 86.5 & \textbf{75.5} \\ \midrule
\multirow{6}{*}{\rotatebox{90}{\textit{Hit}}} & LLaVA-OV-7B & 76.8 & 23.0 & 53.0 & 48.5 & 82.7 & 33.4 & 64.5 & \textbf{83.4} & 4.6 & \textbf{32.0} & 35.8 & 12.6 & 27.0 & 44.4 \\
 & Qwen2VL-7B & 78.4 & 20.6 & 50.4 & 46.1 & 84.7 & 39.1 & 73.4 & 77.1 & 4.0 & 14.7 & 40.0 & 17.4 & 27.5 & 44.1 \\
 & NVILA-8B & 78.2 & \textbf{23.5} & 54.6 & 46.6 & 81.3 & 37.9 & 69.1 & 77.5 & 6.7 & 10.4 & 26.1 & 7.2 & 19.8 & 41.5 \\
 & InternVL2.5-8B & 73.8 & 23.0 & 52.2 & 49.3 & 83.0 & 42.5 & 75.3 & 81.9 & \textbf{9.8} & 19.1 & 34.6 & 19.2 & 27.8 & 45.5 \\
 & VideoLLaMA3-7B & 77.0 & 22.7 & 53.4 & \textbf{51.1} & 83.0 & 40.2 & 67.2 & 79.6 & 4.2 & 7.3 & \textbf{41.5} & 25.1 & \textbf{30.5} & 44.8 \\
 & Qwen2.5VL-7B & \textbf{79.3} & 19.7 & \textbf{56.0} & 49.0 & \textbf{85.6} & \textbf{47.3} & \textbf{81.3} & 82.0 & 9.1 & 19.3 & 40.4 & \textbf{26.5} & 30.3 & \textbf{48.1} \\ \bottomrule
\end{tabular}%
}
\vspace{-5pt}
\end{table}

\begin{table}[!t]
\setlength\tabcolsep{4pt}
\centering
\caption{The accuracy$\uparrow$ (higher is better) of CAPability-QA and the $K\bar{T} \downarrow$ (lower is better) result.}
\label{tab:qa_acc}
\resizebox{0.9\textwidth}{!}{%
\begin{tabular}{@{}llcccccccccccccc@{}}
\toprule
& \textbf{Methods} & \textbf{\begin{tabular}[c]{@{}c@{}}Obj.\\ Cate.\end{tabular}} & \textbf{\begin{tabular}[c]{@{}c@{}}Obj.\\ Num.\end{tabular}} & \textbf{\begin{tabular}[c]{@{}c@{}}Obj.\\ Color\end{tabular}} & \textbf{\begin{tabular}[c]{@{}c@{}}Spa.\\ Rel.\end{tabular}} & \textbf{Scene} & \textbf{\begin{tabular}[c]{@{}c@{}}Cam.\\ Ang.\end{tabular}} & \textbf{OCR} & \textbf{Style} & \textbf{\begin{tabular}[c]{@{}c@{}}Cha.\\ Iden.\end{tabular}} & \textbf{\begin{tabular}[c]{@{}c@{}}\small (D) Obj.\\ Num.\end{tabular}} & \textbf{Act.} & \textbf{\begin{tabular}[c]{@{}c@{}}Cam.\\ Mov.\end{tabular}} & \textbf{Event} & \textbf{Avg.} \\ \midrule
\multirow{7}{*}{\rotatebox{90}{\textit{QA Acc}}} & LLaVA-OV-72B & 95.0 & 54.6 & 63.8 & 94.0 & \textbf{96.2} & 60.6 & 66.3 & 82.0 & 32.1 & \textbf{52.2} & 75.5 & 15.7 & 85.3 & 67.2 \\
& Qwen2VL-72B & 94.7 & 56.1 & 68.6 & 90.7 & 94.0 & 65.0 & 82.4 & 86.6 & 31.3 & 48.9 & 73.0 & 34.1 & 72.7 & 69.1 \\
& InternVL2.5-78B & 95.5 & 56.9 & 67.1 & 90.0 & 91.2 & 54.1 & 79.5 & 82.5 & 19.1 & 49.7 & 79.1 & 23.3 & 81.7 & 66.9 \\
& Qwen2.5VL-72B & 92.7 & \textbf{58.2} & 67.4 & 84.4 & 88.7 & 63.9 & \textbf{87.4} & \textbf{87.3} & 33.4 & 41.4 & 75.8 & \textbf{39.5} & 85.8 & 69.7 \\
& GPT-4o (0806) & 94.5 & 47.2 & 72.5 & 79.5 & 84.5 & \textbf{71.6} & 80.5 & 79.3 & 37.2 & 46.2 & 81.1 & 20.5 & 78.6 & 67.2 \\
& Gemini-1.5-pro & 97.3 & 51.6 & \textbf{78.8} & \textbf{94.4} & 87.1 & 56.8 & 84.8 & 84.2 & 41.2 & 51.2 & 74.4 & 32.2 & 82.8 & \textbf{70.5} \\
& Gemini-2.0-flash & \textbf{98.3} & 46.8 & 73.3 & 93.4 & 95.2 & 57.6 & 84.8 & 74.5 & \textbf{49.1} & 44.2 & \textbf{81.6} & 24.8 & \textbf{86.6} & 70.0 \\ \midrule
\multirow{7}{*}{\rotatebox{90}{$K\bar{T}$}} & LLaVA-OV-72B & 20.3 & 64.3 & 39.6 & 49.6 & 13.9 & 33.0 & 13.7 & 9.4 & 74.8 & 66.1 & 53.2 & 31.4 & 67.1 & 41.3 \\
& Qwen2VL-72B & 16.7 & 62.1 & 37.0 & 46.0 & 12.6 & 35.4 & 10.2 & 10.4 & 84.7 & 78.3 & 47.6 & 53.1 & 60.4 & 42.6 \\
& InternVL2.5-78B & 19.1 & 57.8 & 35.4 & 45.2 & 11.4 & 21.4 & 11.0 & 47.0 & \textbf{8.2} & 73.0 & 62.4 & 68.1 & 69.9 & 40.8 \\
& Qwen2.5VL-72B & 15.3 & 60.7 & 33.5 & 37.2 & 9.1 & 24.3 & 5.9 & 8.5 & 47.4 & 66.2 & 49.3 & 38.2 & 61.1 & 35.1 \\
& GPT-4o (0806) & 13.1 & 55.2 & 26.6 & \textbf{34.7} & \textbf{7.8} & 16.1 & 6.7 & 3.5 & 30.9 & 64.8 & 41.7 & 53.9 & 51.7 & 31.3 \\
& Gemini-1.5-pro & \textbf{11.9} & \textbf{41.0} & \textbf{24.1} & 36.4 & 9.6 & 19.2 & 5.5 & \textbf{3.1} & 18.0 & \textbf{51.6} & \textbf{36.1} & 23.4 & \textbf{47.9} & \textbf{25.2} \\
& Gemini-2.0-flash & 16.3 & 52.6 & 32.8 & 45.1 & 12.6 & \textbf{9.0} & \textbf{4.5} & 3.2 & 25.7 & 58.4 & 45.9 & \textbf{23.1} & 54.5 & 29.5 \\ \bottomrule
\end{tabular}%
}
\vspace{-8pt}
\end{table}

\noindent\textbf{Precision and hit of 7B models.}
The \textit{precision} and \textit{hit} of 7B models are shown in Tab.~\ref{tab:7b_f1score}. Among these 6 models, Qwen2.5VL-7B achieves the best \textit{precision} (75.5\%) and \textit{hit} (48.1\%), demonstrating its awesome ability. Averagely, the 7B models perform a bit worse than 72B models, verifying the scaling law. Among the dimensions, they follow a similar pattern as 72B and closed-source models. Researchers should focus on the thoroughness of object number, object color, spatial relation, style, character identification, and event, and try to improve the ability of the action and the two camera-related dimensions.

\noindent\textbf{QA-based evaluation and the $\bm{K\bar{T}}$ metric.}
Due to the visual element annotation type, we can easily convert our annotations to QA format, evaluate the accuracy of closed-source APIs and 72B models, thus further calculating their $K\bar{T}$ metric, as shown in Tab.~\ref{tab:qa_acc}. We are surprised to find that the difference in their QA accuracy is not significant, which means they can have a similar level of understanding of the correct visual descriptions based on the specific questions. When it comes to the $K\bar{T}$ metric, these models vary a lot. Gemini-1.5-pro performs the best with the lowest $K\bar{T}$ (25.7\%), but the $K\bar{T}$ of LLaVA-OneVision-72B, Qwen2VL-72B, and InternVL2.5-78B comes to more than 40\%, which means they are more likely knowing the answer but cannot tell automatically. This phenomenon shows the performance gap between the strong instruction (QA) task and the weak instruction (caption) task, which may be ignored by previous work.

\subsection{Metric Analysis}
\begin{figure}[!t]
\centering
\includegraphics[width=0.96\textwidth]{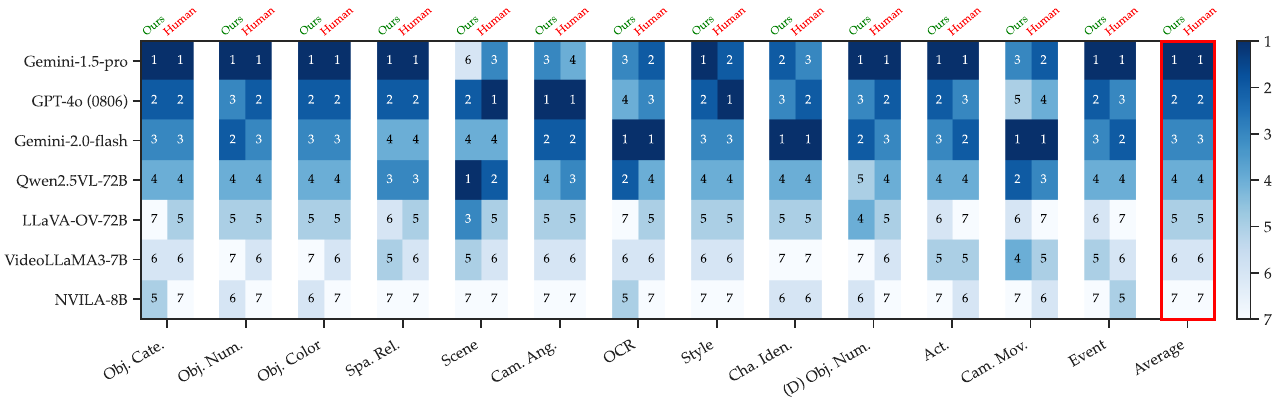}
\vspace{-8pt}
\caption{The metric ranking consistency comparison with human arena evaluation.}
\label{fig:arena_align}
\vspace{-8pt}
\end{figure}

\noindent \textbf{Consistency with human evaluation.}
To demonstrate the reliability of our annotations and evaluation pipeline, we build a human arena to manually compare the caption performance of closed-source models and 72B open-sourced models for each dimension. Specifically, we randomly select 5K pair-wise caption samples for each dimension and require human annotators to select a better one each time. The judging criteria include both the correctness and thoroughness for each dimension. We count the model ranking of each dimension, and compare it with the harmonic mean of \textit{precision} and \textit{hit}. The results are shown in Fig.~\ref{fig:arena_align}. On the scene dimension, scores of all models are high and close in our evaluation, which may cause slight inconsistencies. But for most dimensions, our ranking shows high consistency with human results, leading to high reliability of final evaluation results.

\begin{figure}[!t]
\centering
\includegraphics[width=0.85\textwidth]{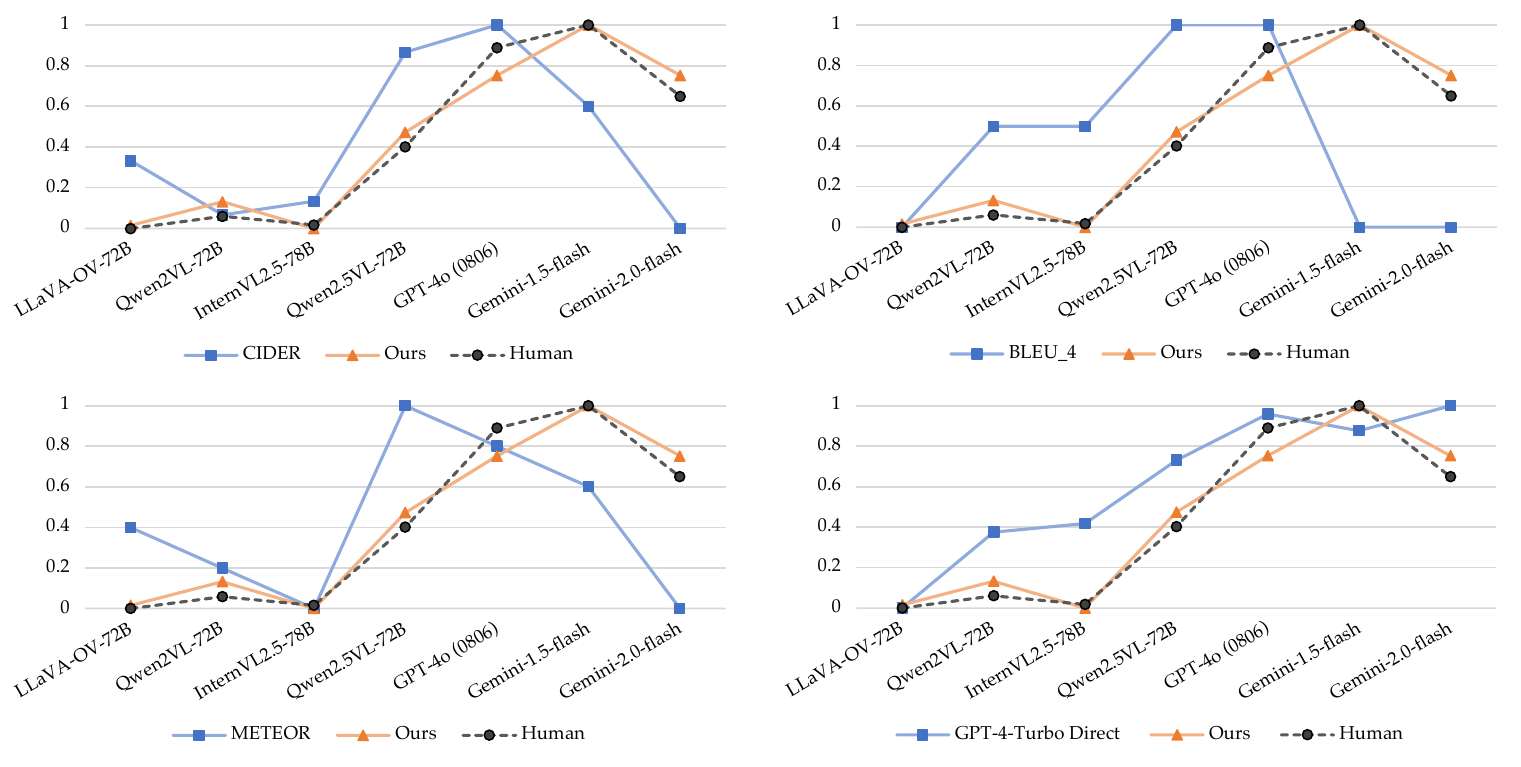}
\vspace{-5pt}
\caption{Consistency comparison with other metrics, we calculate CIDER, BLEU\_4, METEOR metrics, and conduct direct GPT-4-Turbo evaluation. Our evaluation shows the highest consistency.}
\label{fig:traditional_metrics}
\vspace{-10pt}
\end{figure}

\noindent \textbf{Comparison with other metrics.} 
We compare our metrics with other traditional metrics to demonstrate the credibility of our CAPability. We randomly select 200 images or videos for each dimension, and ask human annotators to annotate truthful and detailed captions as the ground truth. Then we calculate traditional metrics (\ie, CIDER~\cite{vedantam2015cider}, BLEU\_4~\cite {papinesi2002bleu}, and METEOR~\cite{banerjee2005meteor}) between the human-annotated and model-generated captions. Besides, we also use GPT-4-Turbo~\cite{gpt4} to directly score the captions from 0 to 5 based on the ground truth. We further randomly select 200 inference samples for each dimension and each model, and manually score them to get human evaluation results as a reference. We report the harmonic mean for CAPability and linearly scale all results to 0-1 for easier comparison. The results are shown in Fig.~\ref{fig:traditional_metrics}. As our result shows high consistency with human evaluation, all traditional metrics fail to align. The phenomenon shows that existing traditional metrics cannot reliably and reasonably evaluate the detailed visual captions. The direct GPT4-Turbo evaluation shows better consistency than traditional metrics, but still has gaps with CAPability, lacks reasonable interpretability, and is too vague for evaluation. Our CAPability can give scores from various aspects with clear judging criteria, thus obtaining an accurate and comprehensive evaluation.

\section{Conclusion}
\label{sec:conclusion}
In this work, we present CAPability, the first comprehensive visual caption benchmark through 6-view and 12-dimensional analysis. Unlike existing benchmarks that rely on oversimplified metrics or limited visual elements, CAPability introduces a correctness-thoroughness dual evaluation framework based on \textit{precision}, \textit{hit}, and $K\bar{T}$. Through this meticulous evaluation process, we uncover specific areas needing improvement across leading models, such as their gap between \textit{precision} and \textit{hit}, the challenges in aspects like camera angle detection, character identification, and action recognition. We also indicate the "know but cannot tell" phenomenon of MLLMs, which may be ignored by previous work. We believe that CAPability will play a pivotal role in advancing the field of visual captioning by encouraging the development of models that holistically understand and describe visual content. 
We open-source all our annotated data to facilitate future research.

\section*{Acknowledgment}
This work is supported by the National Nature Science Foundation of China (62425114, 62121002, U23B2028, 62232006). We thank Alibaba Group for providing human annotation resources, closed-source model APIs support, and GPU computing resources. We thank all outsourcing annotators' efforts, and acknowledge the support of the GPU cluster built by MCC Lab of Information Science and Technology Institution, USTC, and USTC super-computing center for providing computational resources for this project.

%% file: checklist.tex
\newpage
\section*{NeurIPS Paper Checklist}

\begin{enumerate}

\item {\bf Claims}
    \item[] Question: Do the main claims made in the abstract and introduction accurately reflect the paper's contributions and scope?
    \item[] Answer: \answerYes{} 
    \item[] Justification: The main claims presented in the abstract and introduction accurately reflect the contributions and scope of the paper.

\item {\bf Limitations}
    \item[] Question: Does the paper discuss the limitations of the work performed by the authors?
    \item[] Answer: \answerYes{} 
    \item[] Justification: The paper discusses the limitations in the appendix performed by the authors.

\item {\bf Theory assumptions and proofs}
    \item[] Question: For each theoretical result, does the paper provide the full set of assumptions and a complete (and correct) proof?
    \item[] Answer: \answerYes{} 
    \item[] Justification: During the inference process, we set the temperature parameter of the MLLMs to 0 to eliminate randomness and ensure the validity of the theoretical results derived. The proposed metrics are discussed with proper reference in the appendix.

    \item {\bf Experimental result reproducibility}
    \item[] Question: Does the paper fully disclose all the information needed to reproduce the main experimental results of the paper to the extent that it affects the main claims and/or conclusions of the paper (regardless of whether the code and data are provided or not)?
    \item[] Answer: \answerYes{} 
    \item[] Justification: We introduce the detailed steps to obtain experiment results in the paper and explain the implementation details in the appendix. We provide the code and benchmark data to ensure the result can be reproducible. We also provide the inference and evaluation JSON files in the code.

\item {\bf Open access to data and code}
    \item[] Question: Does the paper provide open access to the data and code, with sufficient instructions to faithfully reproduce the main experimental results, as described in supplemental material?
    \item[] Answer: \answerYes{} 
    \item[] Justification: We have provided open-accessed code and data link in the submission.

\item {\bf Experimental setting/details}
    \item[] Question: Does the paper specify all the training and test details (e.g., data splits, hyperparameters, how they were chosen, type of optimizer, etc.) necessary to understand the results?
    \item[] Answer: \answerYes{} 
    \item[] Justification: Detailed specifics are provided in the appendix.

\item {\bf Experiment statistical significance}
    \item[] Question: Does the paper report error bars suitably and correctly defined or other appropriate information about the statistical significance of the experiments?
    \item[] Answer: \answerYes{} 
    \item[] Justification: We report the results of repeating the experiment several times in the appendix to verify the evaluation stability.

\item {\bf Experiments compute resources}
    \item[] Question: For each experiment, does the paper provide sufficient information on the computer resources (type of compute workers, memory, time of execution) needed to reproduce the experiments?
    \item[] Answer: \answerYes{} 
    \item[] Justification: We provide the computing resource we use in the appendix, though it is not the matter we care about for proposing a benchmark.
    
\item {\bf Code of ethics}
    \item[] Question: Does the research conducted in the paper conform, in every respect, with the NeurIPS Code of Ethics \url{https://neurips.cc/public/EthicsGuidelines}?
    \item[] Answer: \answerYes{} 
    \item[] Justification: The research conducted in the paper adheres to the NeurIPS Code of Ethics. All experimental procedures, data handling, and reporting practices were conducted with full compliance to ethical standards, ensuring transparency, integrity, and respect for all stakeholders involved

\item {\bf Broader impacts}
    \item[] Question: Does the paper discuss both potential positive societal impacts and negative societal impacts of the work performed?
    \item[] Answer: \answerYes{} 
    \item[] Justification: We highlight some findings based on our evaluation results, which inspire the researchers. We also discuss the potential positive and negative societal impacts in the appendix.
    
\item {\bf Safeguards}
    \item[] Question: Does the paper describe safeguards that have been put in place for responsible release of data or models that have a high risk for misuse (e.g., pretrained language models, image generators, or scraped datasets)?
    \item[] Answer: \answerYes{} 
    \item[] Justification: As mentioned in the paper, we conduct human filtering when annotating every image and video, ensuring there are no unsafe images or videos released.

\item {\bf Licenses for existing assets}
    \item[] Question: Are the creators or original owners of assets (e.g., code, data, models), used in the paper, properly credited and are the license and terms of use explicitly mentioned and properly respected?
    \item[] Answer: \answerYes{} 
    \item[] Justification: The paper provides appropriate credits and references to the creators or original owners of the assets used. We centrally state the licenses and copyright in the appendix.

\item {\bf New assets}
    \item[] Question: Are new assets introduced in the paper well documented and is the documentation provided alongside the assets?
    \item[] Answer: \answerYes{} 
    \item[] Justification: We have made our benchmark data publicly available, and the link and croissant file are provided in the submission.

\item {\bf Crowdsourcing and research with human subjects}
    \item[] Question: For crowdsourcing experiments and research with human subjects, does the paper include the full text of instructions given to participants and screenshots, if applicable, as well as details about compensation (if any)? 
    \item[] Answer: \answerNA{} 
    \item[] Justification: We ask outsourcing annotators to complete the benchmark construction, but it does not involve crowdsourcing nor research with human subjects.

\item {\bf Institutional review board (IRB) approvals or equivalent for research with human subjects}
    \item[] Question: Does the paper describe potential risks incurred by study participants, whether such risks were disclosed to the subjects, and whether Institutional Review Board (IRB) approvals (or an equivalent approval/review based on the requirements of your country or institution) were obtained?
    \item[] Answer: \answerNA{} 
    \item[] Justification: The paper does not involve crowdsourcing nor research with human subjects.

\item {\bf Declaration of LLM usage}
    \item[] Question: Does the paper describe the usage of LLMs if it is an important, original, or non-standard component of the core methods in this research? Note that if the LLM is used only for writing, editing, or formatting purposes and does not impact the core methodology, scientific rigorousness, or originality of the research, declaration is not required.
    \item[] Answer: \answerYes{} 
    \item[] Justification: We use LLMs for editing typos and formatting, and use LLMs for pre-annotation, followed by human annotation. In our evaluation, we also employ LLM-as-judge evaluation. However, the core motivation, method, and development do not involve LLMs.

\end{enumerate}

%% file: suppl.tex
\appendix
\clearpage

\newcommand\DoToC{%
    \startcontents
    \printcontents{}{1}{\hrulefill\vskip0pt}
    \vskip0pt \noindent\hrulefill
    }

\setcounter{page}{1}
\setcounter{table}{0}
\setcounter{figure}{0}
\setcounter{equation}{0}
\setcounter{footnote}{0}
\renewcommand{\thetable}{A\arabic{table}}
\renewcommand{\thefigure}{A\arabic{figure}}
\renewcommand{\theequation}{A\arabic{equation}}

\begin{center}
    \Large
    \textbf{Appendix}
    \vspace{1.0em}
\end{center}

\noindent\textbf{Overview}
\noindent\DoToC

\section{More Details of CAPability}

\subsection{Details of Dimension Design}
\label{sec:supp_dimension}
We argue that multi-dimensional evaluation is significant to visual caption evaluation and is more comprehensive than previous work. So how to choose proper dimensions? We refer to existing VQA benchmarks~\cite{mmbench, seedbench, mvbench, videomme} and visual generation benchmarks~\cite{geneval, vbench, t2vcompbench}. VQA benchmarks usually design various types of questions to include multi-dimensional evaluation and analysis of MLLMs. For instance, MMBench~\cite{mvbench} defines 20 ability dimensions, including attribute recognition, attribute comparison, action recognition, spatial relationship, physical property, OCR, object localization, image style, image scene, identity reasoning, \textit{etc}. MVBench~\cite{mvbench} covers 20 challenging video tasks including action, object, position, count, scene, pose, attribute, character, cognition, \textit{etc}. Due to the flexible design of questions, VQA benchmarks can be naturally built with comprehensive dimensions. Different from the VQA task, the visual caption task does not require specific questions, but inspects the alignment of visual and textual information. 
Visual generation is the inverse task of visual captioning, as it requires models to generate specific visual content based on detailed textual descriptions. GenEval~\cite{geneval} designs 6 different tasks to evaluate text-to-image alignment, including single object, two object, counting, colors, position, and attribute binding. VBench~\cite{vbench} comprises 16 dimensions, including subject consistency, background consistency, object class, human action, color, spatial relationship, scene, style, \textit{etc}. We follow their explored dimensions to design proper dimensions for visual captioning. Finally, we design 6 views, covering object, global, text, camera, temporal, and knowledge. The object-related view includes object category, object color, object number, and spatial relation, the global-related view includes scene and style, the text-related view evaluates the OCR capability of captions, the camera-related view covers the camera angle and movement, the temporal-related view contains action and event, and we also design a view to evaluate the knowledge of MLLMs, \ie, character identification.

We believe these dimensions contribute to a comprehensive visual caption benchmarking. However, it is undeniable that there may still be other dimensions that also contribute to caption evaluation. This phenomenon exists in all multi-dimensional benchmarks, but the purpose of our design is to find dimensions that are as comprehensive as possible and sufficient to differentiate model capabilities. As we design 12 dimensions, the evaluation is strong enough to evaluate models from various aspects. We also welcome the proposal of more constructive dimensions.

We explain each dimension in detail about what it represents here.
\begin{itemize}
    \item \textbf{Object category.} This dimension measures the ability of whether models can give a correct description of a specific object in the image. The object is randomly selected from the image.
    
    \item \textbf{Object number.} Given a kind of randomly selected object existing in several numbers in an image or a video, this dimension measures the ability of whether models can count the object correctly. For videos, models should watch the whole video and dynamically count the number based on the camera.
    
    \item \textbf{Object color.} Given a kind of randomly selected object in an image, this dimension measures the ability of whether models can correctly describe the color.

    \item \textbf{Spatial relation.} Given two nearby objects in an image, this dimension measures the ability of whether models can correctly describe the spatial relationship of the two objects. We sample 500 images from our collected data, and sample another 500 images from CompreCap~\cite{comprecap}, with their spatial relationship descriptions.
    
    \item \textbf{Scene.} Given an image, this dimension measures the ability of whether models can obtain and tell the global scene of the image correctly.
    
    \item \textbf{Camera angle.} Given an image, this dimension measures the ability of whether models can obtain and tell the camera angle correctly when shooting the image.
    
    \item \textbf{OCR.} Given an image, this dimension measures the ability of whether models can recognize and tell the text appearing in the image correctly.
    
    \item \textbf{Style.} Given an image, this dimension measures the ability of whether models can obtain and tell the global style of the image correctly.
    
    \item \textbf{Character identification.} Given an image, this dimension measures the ability of whether models can recognize the character or the person in the image.
    
    \item \textbf{Action.} Given a video, this dimension measures the ability of whether models can recognize the action in the video. We use the video data of Dream-1K~\cite{dream1k} and re-annotate the action from their annotations.
    
    \item \textbf{Camera movement.} Given a video, this dimension measures the ability of whether models can obtain and tell the camera angle correctly when recording the video. We search videos by ourselves and cut them into short clips, filtering complex movement composition. We only have simple camera movement in our data, but existing models still perform unsatisfactorily.
    
    \item \textbf{Event.} Given a video, this dimension measures the ability of whether models can tell a complete event in the video. We refer Dream-1K~\cite{dream1k} to design this dimension, and we extract the events from their annotations. Different from other dimensions with atom-level elements, the event is usually composed of subjects and actions, which measures the temporal summarization ability of the model.
\end{itemize}

\subsection{Explanation for One Represents All Strategy}
"One represents all" is designed for object selection for object-related dimensions, text for OCR dimension, and action. We further build other dimensions related to attributes and relationships of objects (object color, object number, spatial relation) based on the selected objects. By aggregating results over a large number of samples with random pairing, we achieve statistical coverage across a broad spectrum of relationships, granularities, and contexts. Therefore, our design ensures that both single-object properties and inter-object relationships are robustly evaluated at the benchmark level, while providing a practical and scalable way to balance annotation cost and comprehensive coverage.

We focus on keeping the randomness of element selection, thus covering the whole visual content in a statistical sense, based on the law of large numbers. Therefore, we can get the ability to evaluate the thoroughness of the generated captions by calculating the hit. The details about our efforts to ensure the randomness are as follows.
1) To minimize bias and keep randomness in the pre-annotation stage, we utilize multiple SOTA models (GPT-4o, Gemini-1.5-pro, Qwen-VL-Max) to generate candidate objects and take the union of their outputs. We then use code (Python's random library) to randomly select one object from this union.
2) During manual annotation, as humans tend to select the most obvious objects in the image, annotators are not allowed to reselect or change the target object if the pre-annotation is incorrect, thus mitigating the bias; they are only permitted to judge correctness and filter out incorrect samples. This ensures that neither model nor human bias influences which element is selected for annotation, thus keeping randomness.

\subsection{Details about Human Subjectivity Controlling}
We acknowledge that differences in human annotator judgment can introduce subjectivity, particularly in interpreting synonyms or resolving ambiguous cases. To reduce this, annotators are required to flag any samples for which they have low confidence. All low-confidence samples are then independently annotated by two additional annotators, and the final decision is determined by majority vote among the three annotations. We also have a spot check and verification process. For each dimension, we ask other annotators (or ourselves) to randomly select 20\% of the samples for verification. If the annotation accuracy falls below 97\%, we hold a meeting to highlight the incorrect annotations, revise all annotations for that dimension, and repeat this process until the annotation accuracy meets our requirements.

\begin{figure}[!t]
    \centering
    \includegraphics[width=0.7\linewidth]{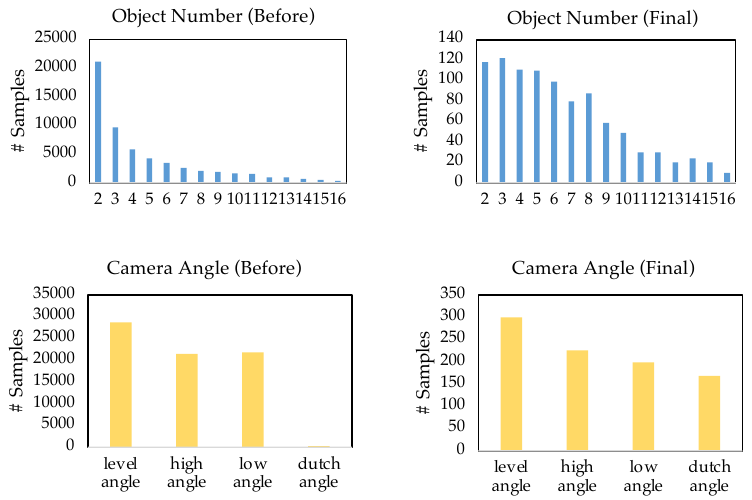}
    \caption{Two examples (object number and camera angle) of data distribution before data balance (pre-annotated) and after data-balanced selection (final human-annotated).}
    \label{fig:data_balance_example}
\end{figure}

\subsection{Details of Data Balance and Final Distribution}
\label{sec:supp_data_balance}
The purpose of data balance is to suppress the long-tail distribution, thus ensuring there are a certain number of samples of different difficulties in the benchmark. Fig.~\ref{fig:data_balance_example} shows two examples of the comparison of pre-annotated data distribution and final human-annotated data distribution. For object number and camera angle dimension, we first randomly sample nearly 75K samples and conduct model pre-annotation. The 75K samples consist of approximately 1/3 SA-1B~\cite{sam}, 1/3 COYO-100M~\cite{coyo700m}, and 1/3 crawled from multiple public datasets and websites. For object number dimension, we select all samples with counting from 2 to 16 same objects within an image. As shown in Fig.~\ref{fig:data_balance_example} (left upper), the counting follows the long-tail distribution, there are fewer images with more objects within an image. Therefore, we conduct data-balanced sampling. Specifically, we separately sample images for different counts, thus forcing the number of each count to be more balanced after the human correcting and filtering process. For camera angle dimension, the dutch angle data is rare, therefore we keep all dutch angle data, and sample the same number of the other three categories. After the human correcting and filtering, the number of these categories varies slightly. The situation in other dimensions is also similar.

As we consider the counts, categories, \textit{etc}. For each dimension to conduct the data balance and not consider the data source (\ie, from SA-1B, COYO-700M, or crawled by ourselves) during this process, the final data source distribution for dimensions of object category, object number, object color, scene, camera angle, OCR varies. 
For data of object-related dimensions (object category, object number, object color, spatial relation), global-related dimensions (scene and style), and a small part of camera angle and OCR dimensions, we collect the base data in a hybrid way. Our approach involves first performing pre-annotation for these dimensions on a large pool $P_{all}$ of images (100K). We then filter out those samples that are not suitable for the corresponding dimension independently, donated as $P^{pre}_i$ (nearly 40K - 80K), where $i$ represents each dimension. After that, we conduct balanced sampling and manual annotation, resulting in approximately 1K samples per dimension, donated as $P^{final}_i$.
For spatial relation, we directly choose 1/2 CompreCap~\cite{comprecap} and 1/2 SA-1B images as SA-1B is more likely to contain high-resolution images with complex object relationship scenes. For style, we choose all realistic images from SA-1B, and crawl animated, special effect, and old-fashioned images by ourselves, all art-related images are from Wikipaintings~\cite{wikipaintings}. For character identification, we use all images from the public dataset, \ie, Wukong-100M~\cite{wukong100m} rather than crawling to ensure proper copyright. For dynamic object number, we directly use data from VSI-Bench~\cite{vsibench}. For action and event, we directly use videos from Dream-1K~\cite{dream1k}. We crawl all videos for camera movement dimension by ourselves, as there is little camera movement data in existing datasets. While there may be some overlap among pre-annotated $P^{pre}_i$, the average of overlapping samples in the final $P^{final}_i$ is 3.95\%, as shown in Tab.~\ref{tab:data_overlap}. We thank all public datasets and benchmarks, their excellent images, videos, and annotations provide much convenience for building our CAPability. 

\begin{table}[!t]
\setlength\tabcolsep{4pt}
\centering
\caption{The overlap among each dimension.}
\label{tab:data_overlap}
\resizebox{0.96\textwidth}{!}{%
\begin{tabular}{@{}cccccccccccc@{}}
\toprule
\textbf{\begin{tabular}[c]{@{}c@{}}Obj.\\ Cate.\end{tabular}} & \textbf{\begin{tabular}[c]{@{}c@{}}Obj.\\ Num.\end{tabular}} & \textbf{\begin{tabular}[c]{@{}c@{}}Obj.\\ Color\end{tabular}} & \textbf{\begin{tabular}[c]{@{}c@{}}Spa.\\ Rel.\end{tabular}} & \textbf{Scene} & \textbf{\begin{tabular}[c]{@{}c@{}}Cam.\\ Ang.\end{tabular}} & \textbf{OCR} & \textbf{Style} & \textbf{\begin{tabular}[c]{@{}c@{}}Cha.\\ Iden.\end{tabular}} & \makebox[0.048\textwidth][c]{\textbf{\small{\begin{tabular}[c]{@{}c@{}}(D) Obj.\\ Num.\end{tabular}}}} & \textbf{\begin{tabular}[c]{@{}c@{}}Cam.\\ Mov.\end{tabular}} & \textbf{Act./Event} \\ \midrule
11.3\% & 10.8\% & 10.5\% & 2.5\% & 4.6\% & 4.6\% & 2.6\% & 0.5\% & 0\% & 0\% & 0\% & 0\% \\ \bottomrule
\end{tabular}%
}
\end{table}

\begin{figure}[!t]
\centering
\includegraphics[width=0.82\textwidth]{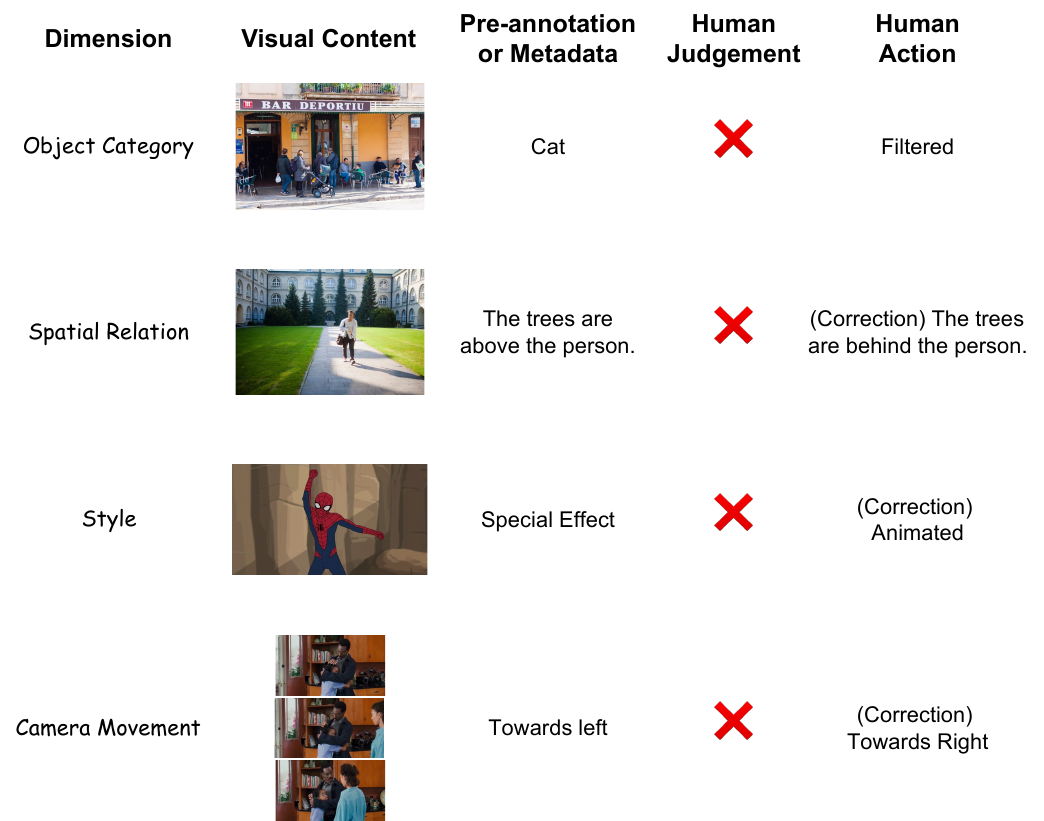}
\caption{Bad case examples of pre-annotation or metadata, human annotators filter or correct the wrong pre-annotation or metadata one by one.}
\label{fig:human_anno_process_example}
\end{figure}

\subsection{Details about Annotators}
We employed 24 outsourcing annotators to complete the annotation task, including 14 female annotators and 10 male annotators. Their ages range from 24 to 30 years old, and all are from mainland China. Each annotator holds at least an associate's or bachelor's degree, demonstrating their expertise level. In total, the annotators labeled 23,824 samples, averaging approximately 993 samples per person. Among these, 3,323 samples were annotated by three different annotators, and the final result was determined through voting.

We specifically instructed annotators on how to handle such cases during the annotation process for each dimension. For example, in the color dimension, if the main object is primarily one color with minor secondary colors (such as a white airplane with some painted markings), only the primary color should be annotated. If the object exhibits multiple primary colors, annotators were asked to label up to three colors if the object can be clearly described within this limit; otherwise, the sample was filtered out. As a result, each object in our dataset is annotated with at most three colors.

\subsection{Ethical Impact}
The geographic and cultural backgrounds of the data are diverse. For data obtained from other public datasets, the distribution of potential bias in our benchmark largely follows that of the original datasets, as we performed random sampling. For the data we collected ourselves, there is a certain degree of cultural bias towards mainland China and East Asian cultures, although mainstream Western cultures are also represented to some extent. However, coverage of cultures from other regions is relatively limited.

\subsection{Discussion about the Designed Metrics}
\label{sec:supp_metrics}
We define two metrics, \textit{precision} and \textit{hit}, to evaluate both the correctness and thoroughness. The \textit{hit} is similar to the meaning of \textit{recall}, which measures how many visual elements in the image/video can be described correctly. However, the \textit{recall} is usually related to the TP/FP/FN/TN system~\cite{allen1955machine}, which is inconsistent with our evaluation situations. To avoid ambiguity and misunderstanding, we name our metric as \textit{hit} rather than \textit{recall}. 

We give the analysis from the TP/FP/FN/TN perspective here. The inconsistency comes from the conflict of the situation definition. In our pipeline, there are three situations when evaluating: 1) MIS, 2) COR, 3) INC, but the TP/FP/FN/TN system does not define the situation of MIS. In the TP/FP/FN/TN system, the \textit{precision} and \textit{recall} are defined as:
\begin{align}
    \label{eq:supp_precision}
    \text{Precision} &= \frac{TP}{TP+FP}, \\
    \label{eq:supp_recall}
    \text{Recall} &= \frac{TP}{TP+FN}.
\end{align}

\begin{figure}[!t]
\centering
\includegraphics[width=0.82\textwidth]{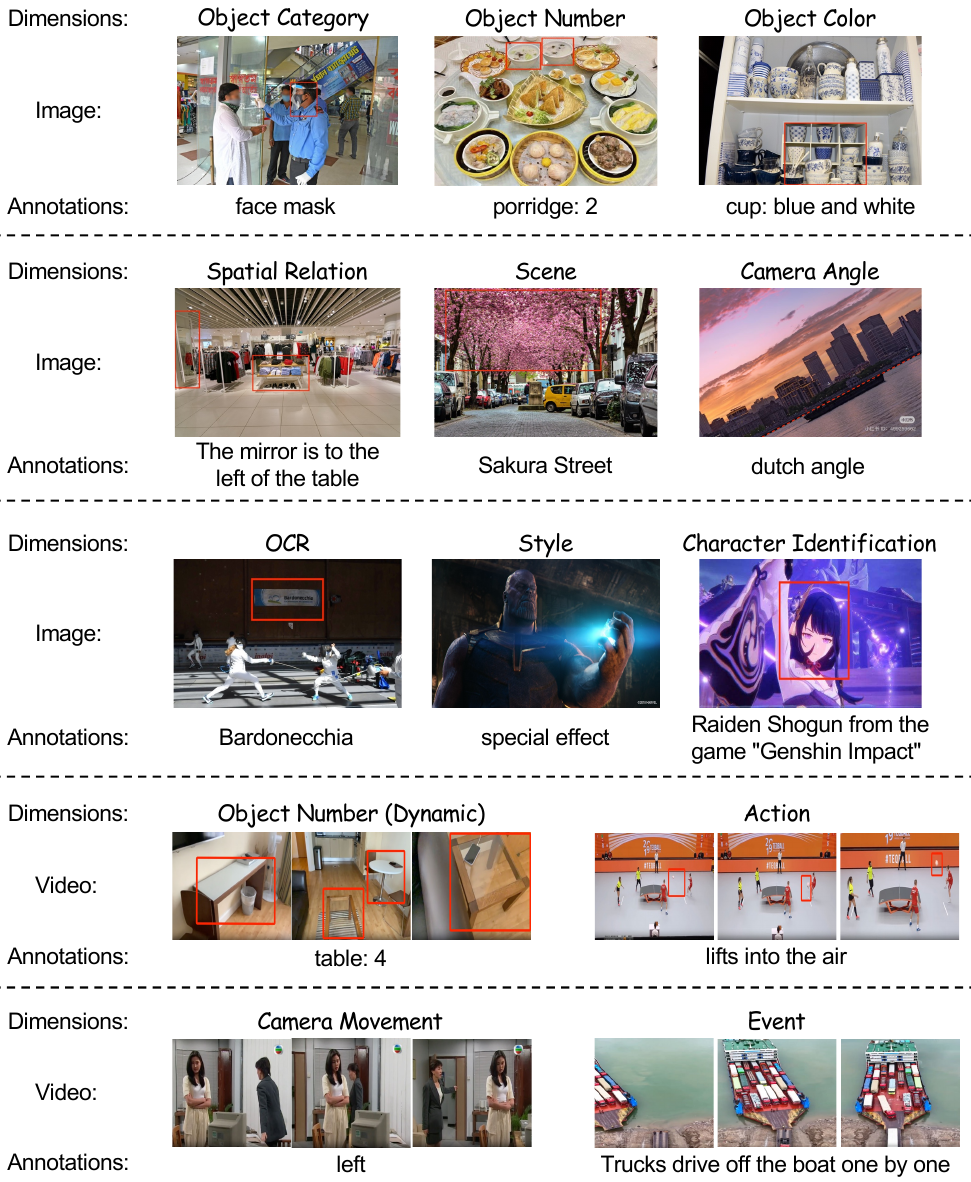}
\caption{Examples of visual content and annotations for each dimension. We outline some visual elements by the red box in the image or video to make them easier to identify.}
\label{fig:anno_example}
\end{figure}

The existence of MIS leads to the ambiguity of the definition of $FN$ and $FP$. If there is no existing MIS, we can only calculate the accuracy without considering \textit{precision} or \textit{recall}. Now we try to analyze with MIS.
Based on the definition of $TP$, we can map our COR to $TP$ with no doubt, as it correctly predicts the answer. If we want to calculate the \textit{precision}, we can map our INC to $FP$, as it wrongly predicts the answer. Therefore, the precision can be calculated by Eq. 1 or Eq.~\ref{eq:supp_precision}, which is consistent.
When we consider \text{recall}, $TP+FN$ should be the number of all ground truths, as there are no negative samples ($TN$) in our annotation, which means $TP+FN = S(\text{COR}) \cup S(\text{INC}) \cup S(\text{MIS})$. This leads to S(INC) being included in both $FP$ and $FN$. As the TP/FP/FN/TN system does not define the MIS, the TP/FP/FN/TN-based \textit{precision} and \textit{recall} cause contradiction. Therefore, we name our metric of $S(\text{COR}) / (S(\text{COR}) \cup S(\text{INC}) \cup S(\text{MIS}))$ as \textit{hit} rather than \textit{recall} to avoid ambiguity as it does not fit the TP/FP/FN/TN-based definition.

In addition to the TP/FP/FN/TN system, there are also other ways to define \textit{precision} and \textit{recall}. MUC-7~\cite{chinchor1998appendix} defines the \textit{precision} and \textit{recall} with the situation of MIS. Apart from COR, INC, MIS, which own the same meaning as ours, MUC-7 also defines SPU, which represents the number of spurious, and equals 0 in our situation. MUC-7 defines the \textit{precision} and \textit{recall} as follows:
\begin{align}
    \text{Precision} &= \frac{S(\text{COR})}{S(\text{COR}) + S(\text{INC}) + S(\text{SPU})}, \\
    \text{Recall} &= \frac{S(\text{COR})}{S(\text{COR}) + S(\text{INC}) + S(\text{MIS})}.
\end{align}
Based on this kind of definition, our \textit{hit} equals the "\textit{recall}". 

\textbf{However, as the TP/FP/FN/TN system is too famous and standard to define the \textit{precision} and \textit{recall}, we finally decide to use the "\textit{hit}" rather than "\textit{recall}" to avoid misunderstanding.}

\begin{figure}[!t]
\centering
\includegraphics[width=0.9\textwidth]{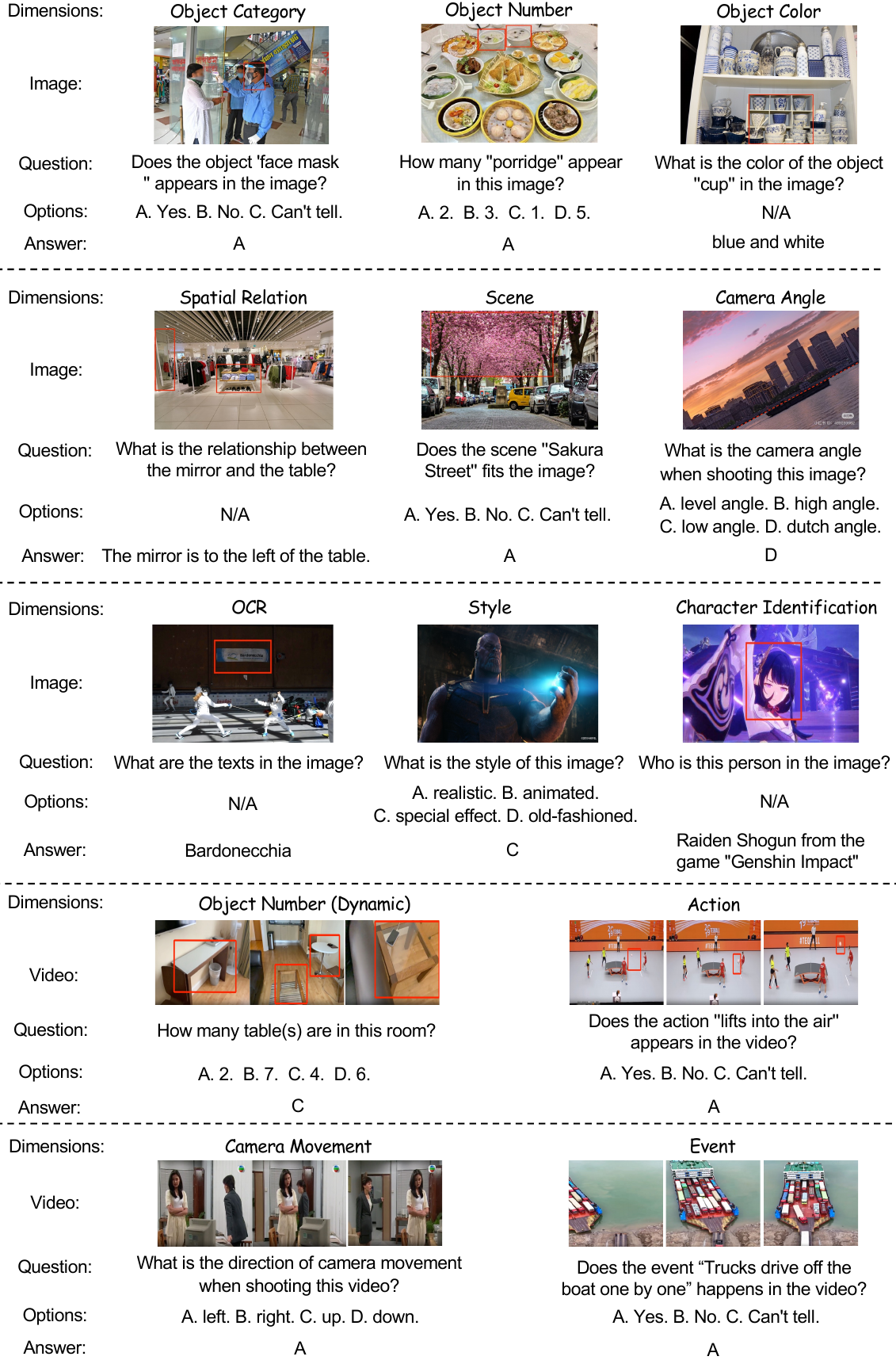}
\caption{Examples of visual content and converted QA annotations for each dimension. The visual content is the same as Fig.~\ref{fig:anno_example}. We outline some visual elements by the red box in the image or video to make them easier to identify.}
\label{fig:qa_anno_example}
\end{figure}

\subsection{Benchmark Examples}
\label{sec:supp_example}

\noindent\textbf{Examples of human annotation process.}
We show some visual cases of the human annotation process in Fig.~\ref{fig:human_anno_process_example}. All of examples are with wrong pre-annotation or metadata. Human annotators check the pre-annotation/metadata one by one and filter/correct the mistakes for each dimension.

\noindent\textbf{Examples of annotations.}
We show some visual cases with our annotations in Fig.~\ref{fig:anno_example}. We outline some visual elements by the red box in the image or video to make them easier to identify. We collect our data from various sources, and we crawled some visual content from the Internet by ourselves, ensuring diversity.

\noindent\textbf{Examples of converted QA pairs.}
As we directly annotate the visual elements in the image or video rather than the caption sentence, we can easily convert our annotation into the format of question-answer (QA) pairs, and we name it as CAPability-QA. We use CAPability-QA to evaluate the QA accuracy and the \textit{know but cannot tell} ($K\bar{T}$) metric. In Fig.~\ref{fig:qa_anno_example}, we also show the same visual cases as Fig.~\ref{fig:anno_example} for each dimension with converted QA format. Most of the dimensions are converted to the format of a multiple-choice QA task with several options, and the object color, OCR, and character identification dimensions are designed as open-ended QA tasks.

\section{More Experimental Analysis}
\subsection{Implementation Details}
\label{sec:supp_impl_details}
We use 4x80G GPUs to run all open-sourced model inference. We use {\tt transformers} to deploy LLaVA-OneVision, InternVL2.5, VideoLLaMA3, and NVILA, use {\tt vLLM} to deploy Qwen2VL and Qwen2.5VL, as their official repositories suggested. 
For all our evaluated model, we follow their official configurations to run the inference. We set the temperature of all open-source models to 0, while keeping the default for closed-source APIs. All maximum output token length is set to 8192. We list the configurations as follows.

\noindent\textbf{LLaVA-OneVision.} We set {\tt anyres-max-9} for image, and uniformly sample 32 frames for video.

\noindent\textbf{Qwen2VL and Qwen2.5VL.} We keep the default minimum and maximum image pixels in package {\tt qwen\_vl\_utils}, which is $4 * 28 * 28$, and $16384 * 28 * 28$, respectively. We also keep default video settings, the fps is set to $2.0$, the maximum frames are $768$, the minimum video pixel is $128 * 28 * 28$, and the maximum video pixel is $768 * 28 * 28$.

\noindent\textbf{InternVL2.5.} We use the official video and image process function and uniformly sample 32 frames for video.

\noindent\textbf{VideoLLaMA3.} We use image model for image dimensions and video model for video dimensions. The fps is set to $1$, and the maximum frames are $180$ for videos.

\noindent\textbf{NVILA.} We use the official image and video process function in {\tt VILA} repository, and uniformly sample 8 frames for videos, as suggested in the official config.

\noindent\textbf{GPT-4o.} Due to the maximum frame number limits of GPT API, we uniformly sample 50 frames for videos, and keep the original spatial size of both images and videos, sending them to the API server.

\noindent\textbf{Gemini-1.5-pro and Gemini-2.0-flash.} As Gemini API supports video, we directly send the original image and video to the API server. For very few videos with too large file size, we downsample the fps to 3, and send the downsampled video to the API server for connection stability.

\begin{figure}[!t]
\centering
\includegraphics[width=0.6\textwidth]{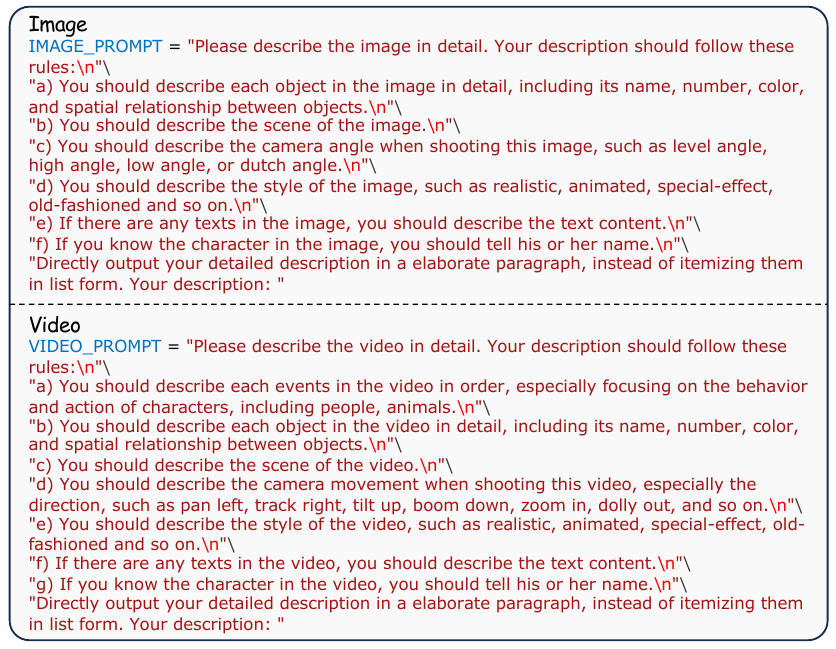}
\caption{The image prompt and video prompt for all models when inferring captions.}
\label{fig:infer_prompt}
\end{figure}

\begin{figure}[!t]
\centering
\includegraphics[width=\textwidth]{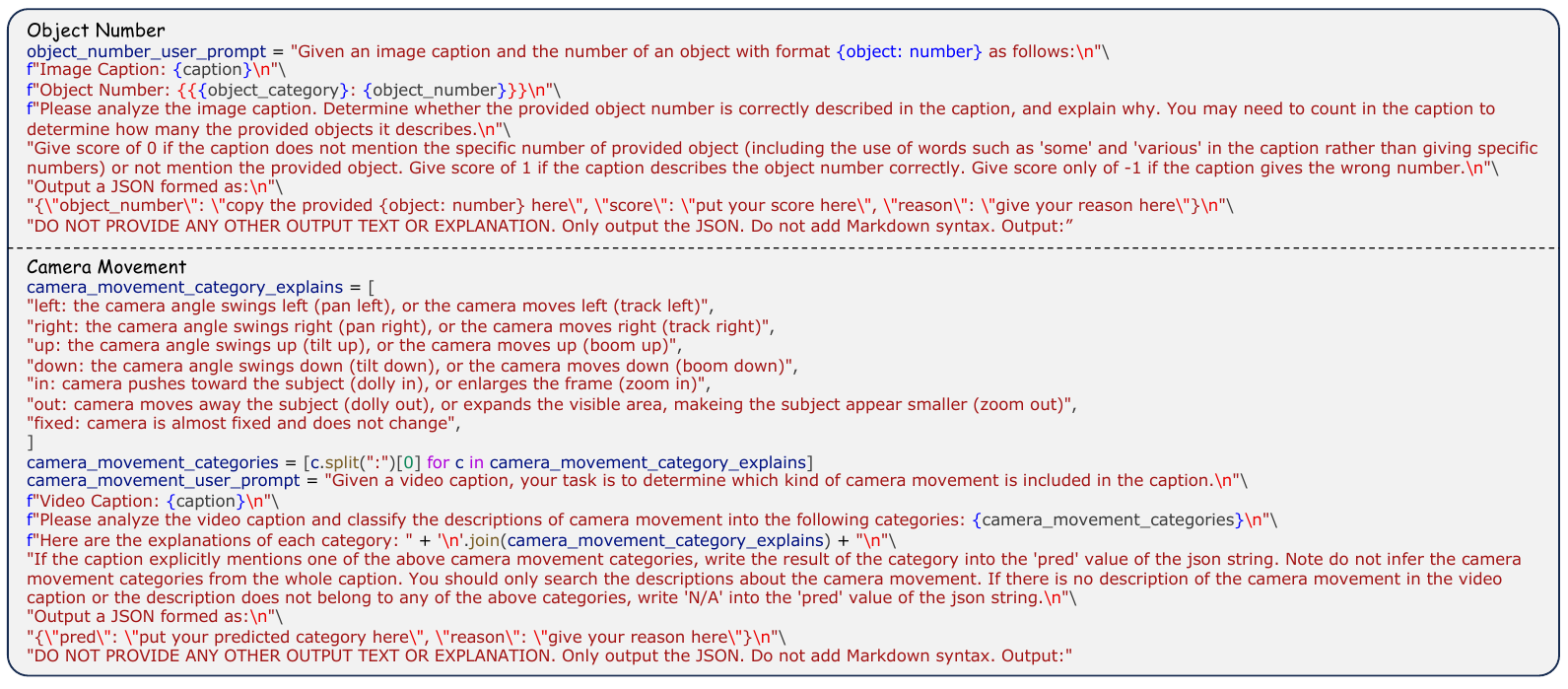}
\caption{Two prompt examples for different types of evaluation sub-tasks. The example of object number represents dimensions with open-ended descriptions, and the example of camera movement represents the dimensions with specific categories.}
\label{fig:eval_prompt}
\end{figure}

\subsection{Prompts of Inference and Evaluation}
\label{sec:supp_prompt}

\noindent\textbf{Inference prompt.}
We use the same prompts for all models to produce the visual captions. The image prompt and video prompt are shown in Fig.~\ref{fig:infer_prompt}. To decrease the inference difficulty, we prompt the models to output the information of all our designed dimensions with a detailed caption. Despite this, the models still show a huge difference in the hit rate of each dimension, which may be due to the variety of training data related to the caption.

\noindent\textbf{Evaluation prompt.}
As we divide the evaluation of dimensions into two types: 1) dimensions with specific categories (\ie, style, camera angle, and camera movement), 2) dimensions with open-ended descriptions. Therefore, we design two kinds of templates for evaluating, and fine-tune them within each dimension. In Fig.~\ref{fig:eval_prompt}, we take the object number dimension and camera movement dimension as examples, to show our prompts for evaluation. For dimensions with specific categories, we ask GPT-4-Turbo to extract the key information and classify the caption into our pre-defined categories or the 'N/A' class. The correct classification is considered positive, the wrong one as negative, and the 'N/A' result is considered a miss. For dimensions with open-ended descriptions, we ask GPT-4-Turbo to directly compare the annotations and the caption, and give out the result of positive, negative, or miss with reasons.

\subsection{More Experimental Results}
\label{sec:supp_exp_result}

\begin{table}[!t]
\setlength\tabcolsep{4pt}
\centering
\caption{The referring ratio of all models, which only reflects the referring ratio of each dimension without considering the accuracy.}
\label{tab:referring_ratio}
\resizebox{0.85\textwidth}{!}{%
\begin{tabular}{@{}lcccccccccccccc@{}}
\toprule
\textbf{Methods} & \textbf{\begin{tabular}[c]{@{}c@{}}Obj.\\ Cate.\end{tabular}} & \textbf{\begin{tabular}[c]{@{}c@{}}Obj.\\ Num.\end{tabular}} & \textbf{\begin{tabular}[c]{@{}c@{}}Obj.\\ Color\end{tabular}} & \textbf{\begin{tabular}[c]{@{}c@{}}Spa.\\ Rel.\end{tabular}} & \textbf{Scene} & \textbf{\begin{tabular}[c]{@{}c@{}}Cam.\\ Ang.\end{tabular}} & \textbf{OCR} & \textbf{Style} & \textbf{\begin{tabular}[c]{@{}c@{}}Cha.\\ Iden.\end{tabular}} & \makebox[0.048\textwidth][c]{\textbf{\small{\begin{tabular}[c]{@{}c@{}}(D) Obj.\\ Num.\end{tabular}}}} & \textbf{Act.} & \textbf{\begin{tabular}[c]{@{}c@{}}Cam.\\ Mov.\end{tabular}} & \textbf{Event} & \textbf{Avg.} \\ \midrule
LLaVA-OV-7B & 96.6 & 33.9 & 60.8 & 54.7 & 86.7 & 79.7 & 73.6 & 98.8 & 5.0 & 34.5 & 92.2 & 62.4 & 30.9 & 62.3 \\
Qwen2VL-7B & 97.6 & 30.2 & 56.8 & 51.8 & 88.7 & 98.0 & 77.0 & 99.9 & 4.8 & 20.0 & 94.0 & 72.0 & 31.7 & 63.3 \\
NVILA-8B & 97.0 & 34.3 & 64.9 & 52.7 & 85.3 & 85.1 & 74.5 & 98.1 & 7.4 & 22.1 & 80.6 & 48.6 & 21.5 & 59.4 \\
InternVL2.5-8B & 97.0 & 38.1 & 60.9 & 55.2 & 87.3 & 100.0 & 84.6 & 100.0 & 20.3 & 22.7 & 91.4 & 94.8 & 32.9 & 68.1 \\
VideoLLaMA3-7B & 95.1 & 34.0 & 62.5 & 56.4 & 85.6 & 93.1 & 74.7 & 98.5 & 5.0 & 9.2 & 96.5 & 83.1 & 34.5 & 63.7 \\
Qwen2.5VL-7B & 96.7 & 26.8 & 63.7 & 55.3 & 89.4 & 99.2 & 86.7 & 100.0 & 11.3 & 20.9 & 92.5 & 98.6 & 35.0 & 67.4 \\
LLaVA-OV-72B & 95.8 & 35.2 & 63.1 & 54.1 & 87.4 & 93.3 & 74.9 & 99.5 & 11.0 & 31.2 & 92.6 & 69.0 & 31.7 & 64.5 \\
Qwen2VL-72B & 97.4 & 35.8 & 64.3 & 56.9 & 89.4 & 99.6 & 83.0 & 100.0 & 6.8 & 21.6 & 93.5 & 77.1 & 34.7 & 66.2 \\
InternVL2.5-78B & 97.2 & 41.7 & 65.3 & 57.0 & 86.7 & 100.0 & 85.5 & 100.0 & 21.3 & 25.7 & 88.2 & 63.3 & 28.8 & 66.2 \\
Qwen2.5VL-72B & 95.6 & 43.3 & 69.2 & 62.3 & 90.7 & 100.0 & 91.8 & 100.0 & 31.4 & 24.4 & 94.8 & 99.4 & 38.9 & 72.5 \\
GPT-4o (0806) & 96.0 & 44.5 & 73.5 & 61.6 & 88.2 & 100.0 & 88.8 & 100.0 & 35.1 & 29.4 & 93.4 & 99.4 & 44.5 & 73.4 \\
Gemini-1.5-pro & 96.1 & 55.3 & 77.0 & 69.0 & 88.1 & 99.9 & 91.4 & 100.0 & 67.5 & 48.9 & 90.5 & 100.0 & 48.6 & 79.4 \\
Gemini-2.0-flash & 96.1 & 39.0 & 67.2 & 58.2 & 87.5 & 100.0 & 93.2 & 99.9 & 46.2 & 30.4 & 92.0 & 99.6 & 44.6 & 73.4 \\ \bottomrule
\end{tabular}%
}
\end{table}

\begin{table}[!t]
\setlength\tabcolsep{3pt}
\centering
\caption{The average metric of image dimensions and video dimensions.}
\label{tab:metric_modality}
\resizebox{\textwidth}{!}{%
\begin{tabular}{lccccccc}
\toprule
\textbf{Models} & \textbf{LLaVA-OV-72B} & \textbf{Qwen2VL-72B} & \textbf{InternVL2.5-78B} & \textbf{Qwen2.5VL-72B} & \textbf{GPT-4o (0806)} & \textbf{Gemini-1.5-pro} & \textbf{Gemini-2.0-flash} \\
\midrule
image precision & 81.4 & 82.2 & 78.2 & 80.7 & 84.4 & 81.1 & 84.6 \\
video precision & 59.5 & 62.9 & 55.6 & 65.0 & 67.6 & 68.7 & 67.3 \\
image hit       & 55.3 & 57.5 & 57.9 & 62.1 & 65.2 & 67.7 & 64.5 \\
video hit       & 26.9 & 29.2 & 22.4 & 33.7 & 36.8 & 43.9 & 37.6 \\
\bottomrule
\end{tabular}
}
\end{table}

\begin{table}[!t]
\setlength\tabcolsep{2pt}
\centering
\caption{The PPL results of each generated caption by different models.}
\label{tab:ppl}
\resizebox{\textwidth}{!}{%
\begin{tabular}{lccccccc}
\toprule
\textbf{PPL models} & \textbf{LLaVA-OV-72B} & \textbf{Qwen2VL-72B} & \textbf{InternVL2.5-78B} & \textbf{Qwen2.5VL-72B} & \textbf{GPT-4o (0806)} & \textbf{Gemini-1.5-pro} & \textbf{Gemini-2.0-flash} \\
\midrule
Qwen3 PPL    & 3.56 & 4.60 & 5.13 & 5.01 & 8.64 & 8.45  & 8.16  \\
LLaMA3.1 PPL & 4.89 & 6.70 & 7.75 & 7.54 & 11.29 & 10.76 & 10.68 \\
\bottomrule
\end{tabular}
}
\end{table}

\noindent\textbf{Referring ratio among all models.}
Apart from the \textit{precision} and \textit{hit}, we can also report another metric, \textit{referring ratio}, which represents the referring ratio about the dimension in visual caption and can be calculated as:
\begin{equation}
    \text{Referring Ratio} = \frac{|S(\text{COR}) \cup S(\text{INC})|}{|S(\text{ALL})|}.
\end{equation}
This metric only considers the pure thoroughness of the caption in each dimension without considering the accuracy.

We report the \textit{referring ratio} in Tab.~\ref{tab:referring_ratio}. For example, it is considered a reference if the caption mentions any object for the object category dimension, or mentions any angle information for the camera angle dimension, but for the object number or color dimension, it is only considered a reference if the caption mentions any number or color information of the correct object. We find the referring ratio seems to increase as the size of models increases, which may be due to more knowledge and stronger instruction following ability for larger models. Among all dimensions, the referring ratio of character identification performs the worst, the existing models prefer to keep silent as they usually cannot recognize the person and character well. The closed-source models would be more likely to reveal the names of characters, and we guess this may be due to stronger knowledge and more diverse training data.

\noindent \textbf{Metric analysis between different modalities.}
Tab.~\ref{tab:metric_modality} shows the metric analysis between image dimensions and video dimensions. Across all models, performance on image dimensions is substantially higher than on video dimensions, for both precision and hit metrics. Even the best-performing models (GPT-4o, Gemini series) show nearly 20\% or greater drop in precision/hit from image to video dimensions. For example, all models perform well on object category, scene, OCR, and style for both precision and hit, but all models cannot achieve a satisfactory level on all video dimensions for hit. This shows that time series modeling and multi-frame information integration are still the main challenges of current MLLMs.
It is true that some dimensions may be inherently more difficult to represent in video than in pictures. This may be due to the following two reasons:
1) There is more redundant content with more visual tokens input into MLLMs, handling the long sequence is likely more difficult than a shorter one.
2) Temporal change and movement should be considered for video dimensions, which may lead to confusion for models to recognize.
Nevertheless, images can also represent the strength of videos in this dimension to a certain extent (such as OCR), because video understanding also requires a good spatial understanding ability as a prerequisite. The video dimensions we designed are more focused on the challenge of temporality. In the future, we will consider introducing more video data to fully evaluate the video's ability in the spatial dimensions.

\begin{figure}[!t]
\centering
\includegraphics[width=\textwidth]{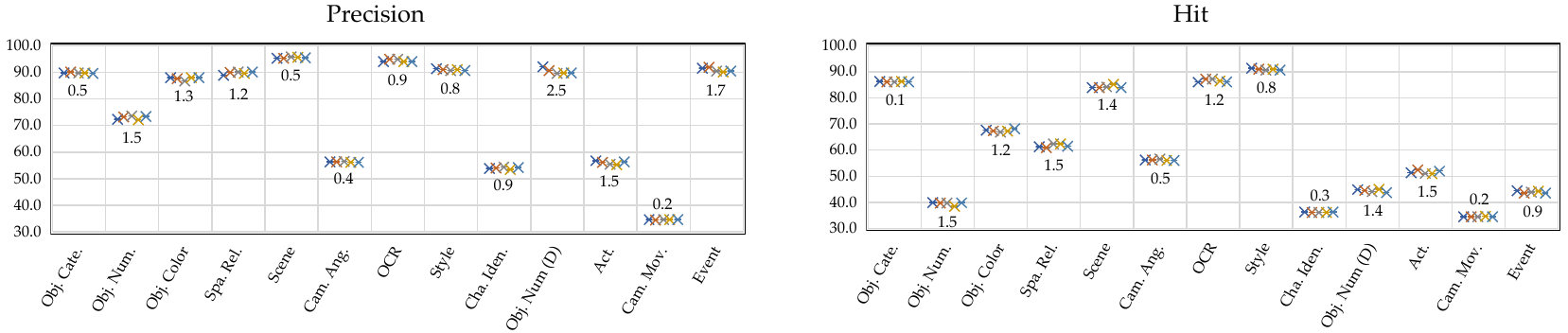}
\caption{The evaluation of repeating 5 times for Gemini-1.5-pro captions. We tag the fluctuation range beside the data point.}
\label{fig:stability1}
\end{figure}

\begin{figure}[!t]
\centering
\includegraphics[width=\textwidth]{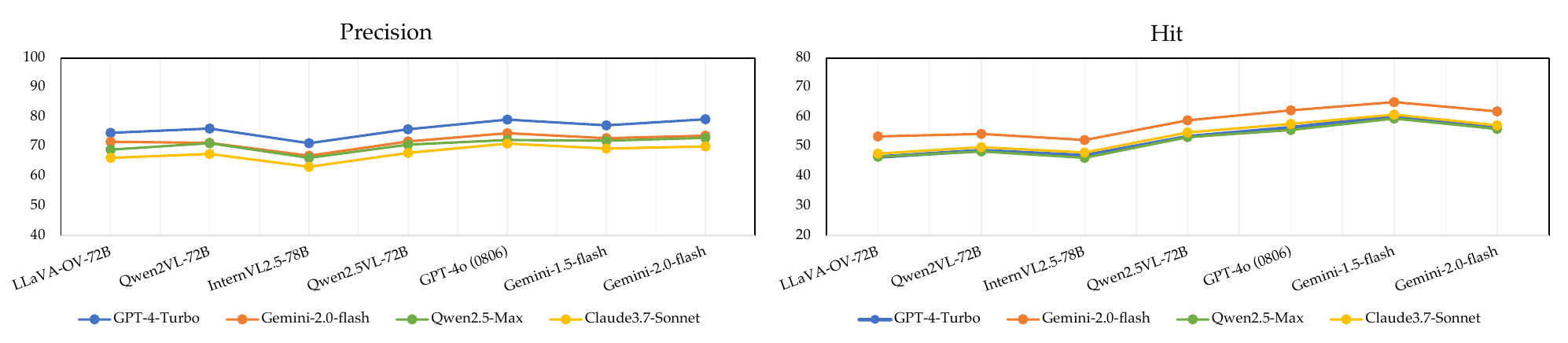}
\caption{The stability analysis with three different evaluation models on 7 MLLMs' captions. The results on all metrics show a high degree of consistency.}
\label{fig:stability2}
\end{figure}

\noindent \textbf{The naturalness and coherence analysis.}
In the current landscape of MLLMs, the naturalness and coherence of generated captions are generally no longer a sufficient or even primary differentiator of model performance. Linguistic quality is more considered for NLP language models, and modern LLMs consistently generate grammatically correct and fluent sentences~\cite{chang2024survey}. As MLLMs are built from LLMs, the main challenge and research focus of MLLMs has shifted toward evaluating the factual accuracy and relevance of the generated captions, rather than their linguistic quality. For example, the papers of MLLMs do not report any metrics about the naturalness and coherence, such as PPL~\cite{llavaov, qwen2.5vl}, modern benchmarks for MLLMs do not measure the naturalness and coherence either~\cite{mmmu, videomme}.
To support the view that the naturalness and coherence of generated captions are not the primary challenges, we also further calculate perplexity (PPL) of the generated captions of each model. PPL is a fundamental metric for language models that gauges how “surprised” the model is by a given text, thus reflecting their naturalness and fluency, the lower PPL means better coherence.
Specifically, we conduct qwen3-32B and LLaMA-3.1-8B-Instruct to calculate the PPL of generated captions from these models, as shown in Tab.~\ref{tab:ppl}. (There seems to be some gap between the closed-source APIs and open-source models. This may be because the training data distribution among open-source models is more similar to Qwen3 and LLaMA3.1 than closed-source APIs.) Among closed-source APIs, the PPLs of them are similar to or lower than GPT-4o. Among open-source models, the PPLs of them are similar to or lower than Qwen2.5VL-72B. Based on the common sense of GPT-4o and Qwen2.5VL-72B can generate coherent and human-like sentences, we can draw a conclusion that the naturalness and coherence are not the main challenge for all these models.

\noindent \textbf{Evaluation stability.}
To validate the stability and robustness of our GPT-4-Turbo-based evaluation method, we take the inferred caption of Gemini-1.5-pro as the example, run our evaluation 5 times, and the result is shown in Fig.~\ref{fig:stability1}. We tag the fluctuation range, \ie, the difference between the maximum and minimum scores, besides the data point. Fig.~\ref{fig:stability1} shows our strong stability, and our average range of precision and recall are 1.1\% and 1.0\%, respectively. This demonstrates the reliability and interpretability of our evaluation method, which matches annotated elements in the generated captions.
Moreover, we introduce three other models, Gemini-2.0-flash, Qwen2.5-Max, and Claude3.7-Sonnet to replace the GPT-4-Turbo in our evaluation pipeline. All of them are the most popular and powerful SOTA language models. We re-run the evaluation process with new evaluation models with the same evaluation prompts. The average result of evaluations for closed-source models and open-source 72B models is shown in Fig.~\ref{fig:stability2}. Though the judgment criteria for each model may be slightly different, leading to only a slight difference in the scores, the evaluation results show high consistency as they keep the same rank of these MLLMs as GPT-4-Turbo. This further demonstrates the high reliability, interpretability, and stability of our evaluation methods.

\begin{figure}[!t]
\centering
\includegraphics[width=0.88\textwidth]{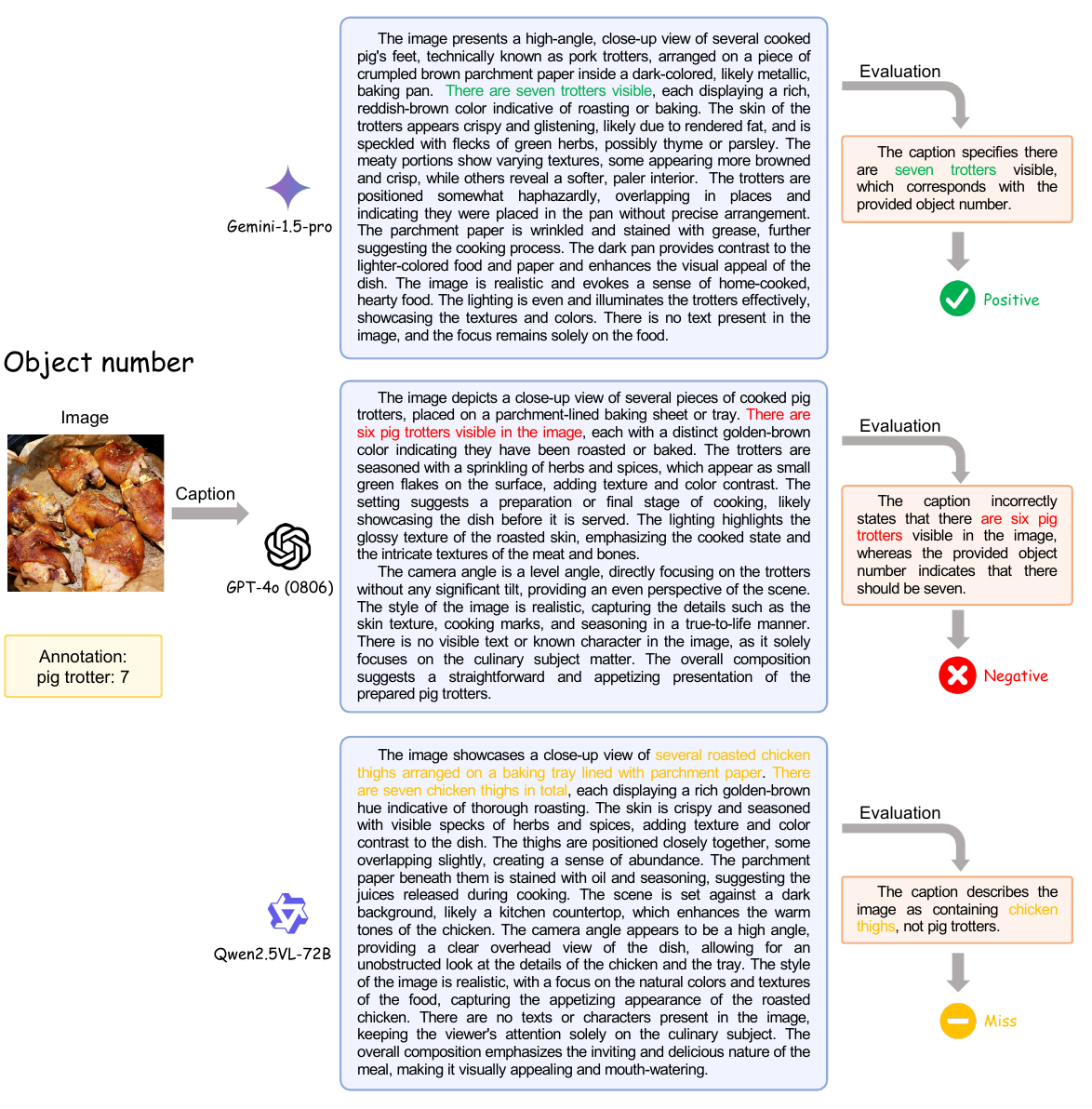}
\caption{Examples of inference and evaluation on object number dimension. We select the inferred caption from Gemini-1.5-pro, GPT-4o, and Qwen2.5VL-72B as instances.}
\label{fig:case_obj_num}
\end{figure}

\begin{figure}[!t]
\centering
\includegraphics[width=0.88\textwidth]{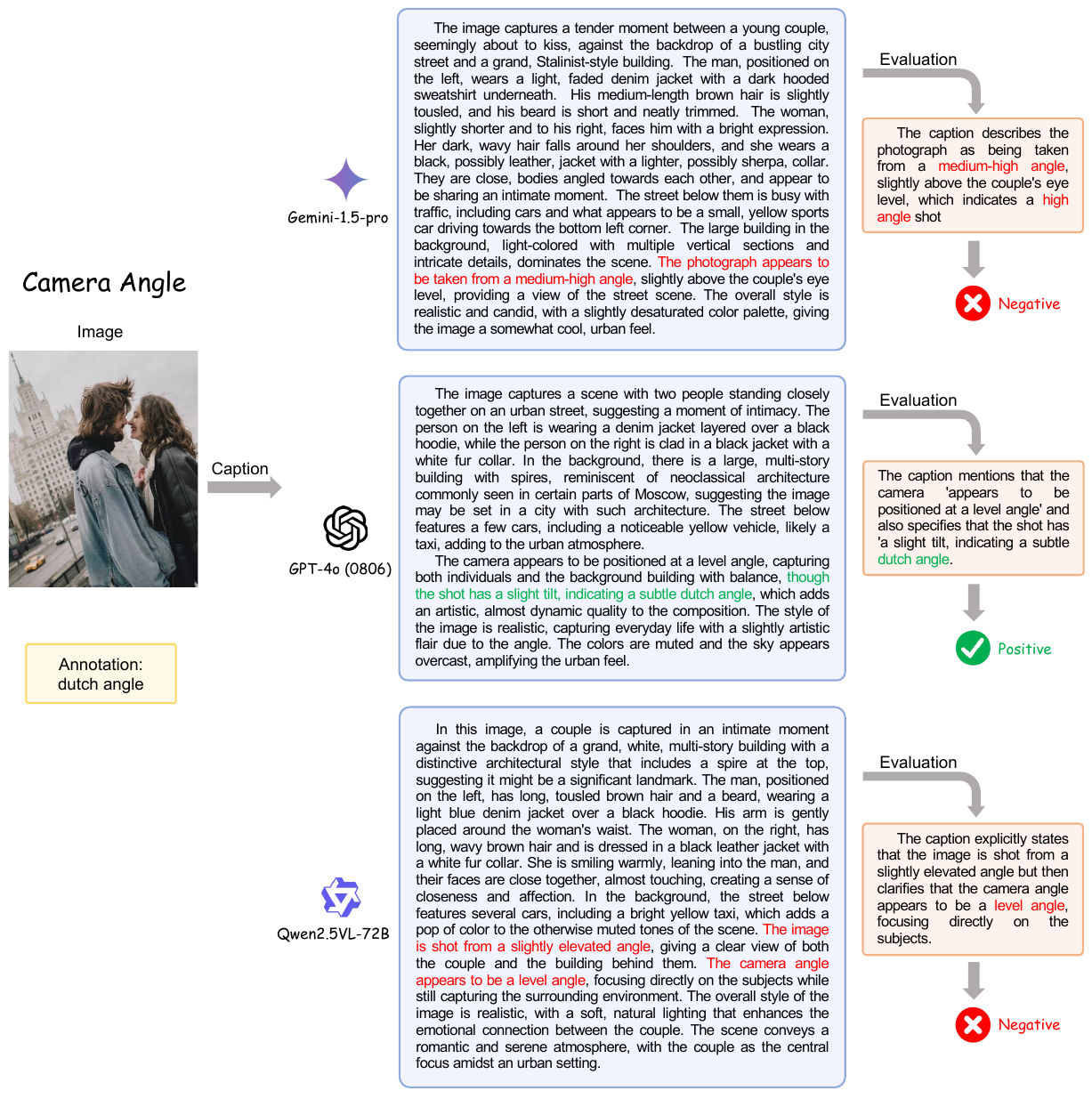}
\caption{Examples of inference and evaluation on camera angle dimension. We select the inferred caption from Gemini-1.5-pro, GPT-4o, and Qwen2.5VL-72B as instances.}
\label{fig:case_cam_ang}
\end{figure}

\subsection{Visualization of Inference and Evaluation.}
\label{sec:supp_vis}
In Fig.~\ref{fig:case_obj_num} and Fig.~\ref{fig:case_cam_ang}, we visualize the inferred caption of Gemini-1.5-pro, GPT-4o (0806), and Qwen2.5VL-72B in object number dimension and camera angle dimension. In Fig.~\ref{fig:case_obj_num}, the annotation of the given image is 7 pig trotters. Gemini-1.5-pro refers to the correct number of pig trotters, and we thus give it a positive. GPT-4o recognizes the trotters in the image, but counts with a wrong number, 6, and we thus give it a negative. As for Qwen2.5VL-72B, it says there are 7 chicken thighs in the image, recognizing the wrong object category. However, this is the dimension of the object number, and therefore we only evaluate the correctness and thoroughness of the number, without considering the categories of objects. As Qwen2.5VL-72B does not mention the pig trotters, we give it a miss. In Fig.~\ref{fig:case_cam_ang}, the annotation of the given image is dutch angle, which means the lens has a certain angle of deflection along the central axis, making the horizon crooked. Gemini-1.5-pro says the image is taken from a medium-high angle, and we classify it into the high angle category, thus negative. GPT-4o explicitly points it out as a subtle dutch angle, thus is classified into the dutch angle category, which is positive. Qwen2.5VL-72B describes the image shot from a slightly elevated angle, and it appears to be a level angle, which is also negative. These two figures show our evaluation pipeline, which is precise and reliable.

\section{Copyright and License}
CAPability comprises data from SA-1B, COYO-700M, Wukong-100M, Wikipaintings, VSI-Bench, CompreCap, Dream-1K, and uses some craweled data from inner retrieval system, each governed by its own licensing terms:
\begin{itemize}
    \item \textbf{SA-1B}: SA-1B Dataset Research License\footnote{https://ai.meta.com/datasets/segment-anything-downloads/}
    \item \textbf{COYO-100M}: CC-BY-4.0 License
    \item \textbf{Wukong-100M}: CC-BY-NC-SA-4.0 License
    \item \textbf{Wikipaintings}: BSD 2-clause License
    \item \textbf{VSI-Bench}: Apache-2.0 License
    \item \textbf{CompreCap}: CC-BY-4.0 License
    \item \textbf{Dream-1K}: Apache-2.0 License
    \item \textbf{Craweled data}: Our craweled data are all retrieved from inner multi-modal retrieval system, which contains various public datasets, and visual contents from websites with CC0 license.
    
\end{itemize}

\section{Limitations}
Different from multi-choice VQA benchmarks which evaluate models by definite and explicit choice accuracy, our CAPability is a visual caption benchmark, which still depends on LLMs for evaluation. Therefore, it is still hard to ensure a completely correct evaluation. We try to split the evaluation into several dimensions, thus makes the evaluation as simple and clear as possible. Therefore, the LLM-based evaluation can be more accurate. Due to the LLM language capability limitation, it can still make wrong judgments and requires a carefully designed prompt for constraint.

\section{Societal Impacts}
As our proposed CAPability can perform comprehensive caption evaluations of MLLMs, this work can help LLM users make informed choice, and leads the community to build more and more strong MLLMs. The potential negative impacts are similar to other LLM-related works, The development of MLLMs and MLLMs' benchmarks pose societal risks like the perpetuation of biases, the potential for misinformation, job displacement, \textit{etc}.

%% file: main.bib
@String(CVPR= {IEEE Conf. Comput. Vis. Pattern Recog.})

@String(AAAI = {AAAI})

@String(CVPR  = {CVPR})

@inproceedings{fang2015captions,
  title={From captions to visual concepts and back},
  author={Fang, Hao and Gupta, Saurabh and Iandola, Forrest and Srivastava, Rupesh K and Deng, Li and Doll{\'a}r, Piotr and Gao, Jianfeng and He, Xiaodong and Mitchell, Margaret and Platt, John C and others},
  booktitle={Proceedings of the IEEE conference on computer vision and pattern recognition},
  pages={1473--1482},
  year={2015}
}

@article{xu2015show,
  title={Show, attend and tell: Neural image caption generation with visual attention},
  author={Xu, Kelvin},
  journal={arXiv preprint arXiv:1502.03044},
  year={2015}
}

@article{mscoco,
  title={Microsoft coco captions: Data collection and evaluation server},
  author={Chen, Xinlei and Fang, Hao and Lin, Tsung-Yi and Vedantam, Ramakrishna and Gupta, Saurabh and Doll{\'a}r, Piotr and Zitnick, C Lawrence},
  journal={arXiv preprint arXiv:1504.00325},
  year={2015}
}

@inproceedings{nocaps,
  title={Nocaps: Novel object captioning at scale},
  author={Agrawal, Harsh and Desai, Karan and Wang, Yufei and Chen, Xinlei and Jain, Rishabh and Johnson, Mark and Batra, Dhruv and Parikh, Devi and Lee, Stefan and Anderson, Peter},
  booktitle={Proceedings of the IEEE/CVF international conference on computer vision},
  pages={8948--8957},
  year={2019}
}

@inproceedings{msrvtt,
  title={Msr-vtt: A large video description dataset for bridging video and language},
  author={Xu, Jun and Mei, Tao and Yao, Ting and Rui, Yong},
  booktitle={Proceedings of the IEEE conference on computer vision and pattern recognition},
  pages={5288--5296},
  year={2016}
}

@inproceedings{vatex,
  title={Vatex: A large-scale, high-quality multilingual dataset for video-and-language research},
  author={Wang, Xin and Wu, Jiawei and Chen, Junkun and Li, Lei and Wang, Yuan-Fang and Wang, William Yang},
  booktitle={Proceedings of the IEEE/CVF international conference on computer vision},
  pages={4581--4591},
  year={2019}
}

@inproceedings{papinesi2002bleu,
  title={Bleu: A method for automatic evaluation of machine translation},
  author={Papinesi, K},
  booktitle={Proc. 40th Actual Meeting of the Association for Computational Linguistics (ACL), 2002},
  pages={311--318},
  year={2002}
}

@inproceedings{vedantam2015cider,
  title={Cider: Consensus-based image description evaluation},
  author={Vedantam, Ramakrishna and Lawrence Zitnick, C and Parikh, Devi},
  booktitle={Proceedings of the IEEE conference on computer vision and pattern recognition},
  pages={4566--4575},
  year={2015}
}

@inproceedings{banerjee2005meteor,
  title={METEOR: An automatic metric for MT evaluation with improved correlation with human judgments},
  author={Banerjee, Satanjeev and Lavie, Alon},
  booktitle={Proceedings of the acl workshop on intrinsic and extrinsic evaluation measures for machine translation and/or summarization},
  pages={65--72},
  year={2005}
}

@article{detailcaps,
  title={Benchmarking and Improving Detail Image Caption},
  author={Dong, Hongyuan and Li, Jiawen and Wu, Bohong and Wang, Jiacong and Zhang, Yuan and Guo, Haoyuan},
  journal={arXiv preprint arXiv:2405.19092},
  year={2024}
}

@article{comprecap,
  title={Benchmarking Large Vision-Language Models via Directed Scene Graph for Comprehensive Image Captioning},
  author={Lu, Fan and Wu, Wei and Zheng, Kecheng and Ma, Shuailei and Gong, Biao and Liu, Jiawei and Zhai, Wei and Cao, Yang and Shen, Yujun and Zha, Zheng-Jun},
  journal={arXiv preprint arXiv:2412.08614},
  year={2024}
}

@article{dream1k,
  title={Tarsier: Recipes for training and evaluating large video description models},
  author={Wang, Jiawei and Yuan, Liping and Zhang, Yuchen and Sun, Haomiao},
  journal={arXiv preprint arXiv:2407.00634},
  year={2024}
}

@article{auroracap,
  title={AuroraCap: Efficient, Performant Video Detailed Captioning and a New Benchmark},
  author={Chai, Wenhao and Song, Enxin and Du, Yilun and Meng, Chenlin and Madhavan, Vashisht and Bar-Tal, Omer and Hwang, Jeng-Neng and Xie, Saining and Manning, Christopher D},
  journal={arXiv preprint arXiv:2410.03051},
  year={2024}
}

@article{llava,
  title={Visual instruction tuning},
  author={Liu, Haotian and Li, Chunyuan and Wu, Qingyang and Lee, Yong Jae},
  journal={NeurIPS},
  volume={36},
  year={2023}
}

@inproceedings{llava1.5,
  title={Improved baselines with visual instruction tuning},
  author={Liu, Haotian and Li, Chunyuan and Li, Yuheng and Lee, Yong Jae},
  booktitle={CVPR},
  pages={26296--26306},
  year={2024}
}

@misc{llavanext,
    title={LLaVA-NeXT: Improved reasoning, OCR, and world knowledge},
    howpublished={\url{https://llava-vl.github.io/blog/2024-01-30-llava-next/}},
    author={Liu, Haotian and Li, Chunyuan and Li, Yuheng and Li, Bo and Zhang, Yuanhan and Shen, Sheng and Lee, Yong Jae},
    month={January},
    year={2024}
}

@misc{llavanextvideo,
  title={LLaVA-NeXT: A Strong Zero-shot Video Understanding Model},
  howpublished={\url{https://llava-vl.github.io/blog/2024-04-30-llava-next-video/}},
  author={Zhang, Yuanhan and Li, Bo and Liu, haotian and Lee, Yong jae and Gui, Liangke and Fu, Di and Feng, Jiashi and Liu, Ziwei and Li, Chunyuan},
  month={April},
  year={2024}
}

@article{llavaov,
  title={Llava-onevision: Easy visual task transfer},
  author={Li, Bo and Zhang, Yuanhan and Guo, Dong and Zhang, Renrui and Li, Feng and Zhang, Hao and Zhang, Kaichen and Li, Yanwei and Liu, Ziwei and Li, Chunyuan},
  journal={arXiv preprint arXiv:2408.03326},
  year={2024}
}

@article{llavavideo,
  title={Video Instruction Tuning With Synthetic Data},
  author={Zhang, Yuanhan and Wu, Jinming and Li, Wei and Li, Bo and Ma, Zejun and Liu, Ziwei and Li, Chunyuan},
  journal={arXiv preprint arXiv:2410.02713},
  year={2024}
}

@article{minigpt4,
  title={Minigpt-4: Enhancing vision-language understanding with advanced large language models},
  author={Zhu, Deyao and Chen, Jun and Shen, Xiaoqian and Li, Xiang and Elhoseiny, Mohamed},
  journal={arXiv preprint arXiv:2304.10592},
  year={2023}
}

@inproceedings{internvl,
  title={Internvl: Scaling up vision foundation models and aligning for generic visual-linguistic tasks},
  author={Chen, Zhe and Wu, Jiannan and Wang, Wenhai and Su, Weijie and Chen, Guo and Xing, Sen and Zhong, Muyan and Zhang, Qinglong and Zhu, Xizhou and Lu, Lewei and others},
  booktitle={Proceedings of the IEEE/CVF Conference on Computer Vision and Pattern Recognition},
  pages={24185--24198},
  year={2024}
}

@article{internvl2.5,
  title={Expanding performance boundaries of open-source multimodal models with model, data, and test-time scaling},
  author={Chen, Zhe and Wang, Weiyun and Cao, Yue and Liu, Yangzhou and Gao, Zhangwei and Cui, Erfei and Zhu, Jinguo and Ye, Shenglong and Tian, Hao and Liu, Zhaoyang and others},
  journal={arXiv preprint arXiv:2412.05271},
  year={2024}
}

@article{qwen2.5,
  title={Qwen2.5 technical report},
  author={Yang, An and Yang, Baosong and Zhang, Beichen and Hui, Binyuan and Zheng, Bo and Yu, Bowen and Li, Chengyuan and Liu, Dayiheng and Huang, Fei and Wei, Haoran and others},
  journal={arXiv preprint arXiv:2412.15115},
  year={2024}
}

@article{qwenvl,
  title={Qwen-vl: A frontier large vision-language model with versatile abilities},
  author={Bai, Jinze and Bai, Shuai and Yang, Shusheng and Wang, Shijie and Tan, Sinan and Wang, Peng and Lin, Junyang and Zhou, Chang and Zhou, Jingren},
  journal={arXiv preprint arXiv:2308.12966},
  year={2023}
}

@article{qwen2vl,
  title={Qwen2-VL: Enhancing Vision-Language Model's Perception of the World at Any Resolution},
  author={Wang, Peng and Bai, Shuai and Tan, Sinan and Wang, Shijie and Fan, Zhihao and Bai, Jinze and Chen, Keqin and Liu, Xuejing and Wang, Jialin and Ge, Wenbin and others},
  journal={arXiv preprint arXiv:2409.12191},
  year={2024}
}

@misc{qwen2.5vl,
    title = {Qwen2.5-VL},
    howpublished={\url{https://qwenlm.github.io/blog/qwen2.5-vl/}},
    author = {Qwen Team},
    month = {January},
    year = {2025}
}

@article{nvila,
  title={NVILA: Efficient frontier visual language models},
  author={Liu, Zhijian and Zhu, Ligeng and Shi, Baifeng and Zhang, Zhuoyang and Lou, Yuming and Yang, Shang and Xi, Haocheng and Cao, Shiyi and Gu, Yuxian and Li, Dacheng and others},
  journal={arXiv preprint arXiv:2412.04468},
  year={2024}
}

@article{videollama3,
  title={VideoLLaMA 3: Frontier Multimodal Foundation Models for Image and Video Understanding},
  author={Zhang, Boqiang and Li, Kehan and Cheng, Zesen and Hu, Zhiqiang and Yuan, Yuqian and Chen, Guanzheng and Leng, Sicong and Jiang, Yuming and Zhang, Hang and Li, Xin and others},
  journal={arXiv preprint arXiv:2501.13106},
  year={2025}
}

@article{gpt3,
  title={Language models are few-shot learners},
  author={Brown, Tom and Mann, Benjamin and Ryder, Nick and Subbiah, Melanie and Kaplan, Jared D and Dhariwal, Prafulla and Neelakantan, Arvind and Shyam, Pranav and Sastry, Girish and Askell, Amanda and others},
  journal={Advances in neural information processing systems},
  volume={33},
  pages={1877--1901},
  year={2020}
}

@article{gpt4,
  title={Gpt-4 technical report},
  author={Achiam, Josh and Adler, Steven and Agarwal, Sandhini and Ahmad, Lama and Akkaya, Ilge and Aleman, Florencia Leoni and Almeida, Diogo and Altenschmidt, Janko and Altman, Sam and Anadkat, Shyamal and others},
  journal={arXiv preprint arXiv:2303.08774},
  year={2023}
}

@misc{gpt4v,
  title = {Gpt-4V(ision) system card},
  author = {OpenAI},
  year = {2023},
  howpublished={\url{https://cdn.openai.com/papers/GPTV_System_Card.pdf}}
}

@misc{gpt4o,
  title = {Gpt-4O(mini) system card},
  author = {OpenAI},
  year = {2024},
  howpublished={\url{https://openai.com/index/hello-gpt-4o/}}
}

@article{gemini1.5,
  title={Gemini 1.5: Unlocking multimodal understanding across millions of tokens of context},
  author={Reid, Machel and Savinov, Nikolay and Teplyashin, Denis and Lepikhin, Dmitry and Lillicrap, Timothy and Alayrac, Jean-baptiste and Soricut, Radu and Lazaridou, Angeliki and Firat, Orhan and Schrittwieser, Julian and others},
  journal={arXiv preprint arXiv:2403.05530},
  year={2024}
}

@misc{gemini2.0,
    title = {Gemini 2.0 is now available to everyone},
    howpublished={\url{https://blog.google/technology/google-deepmind/gemini-model-updates-february-2025/}},
    author = {Google Deepmind},
    month = {February},
    year = {2025}
}

@misc{vicuna,
    title = {Vicuna: An Open-Source Chatbot Impressing GPT-4 with 90\%* ChatGPT Quality},
    howpublished={\url{https://lmsys.org/blog/2023-03-30-vicuna/}},
    author = {Chiang, Wei-Lin and Li, Zhuohan and Lin, Zi and Sheng, Ying and Wu, Zhanghao and Zhang, Hao and Zheng, Lianmin and Zhuang, Siyuan and Zhuang, Yonghao and Gonzalez, Joseph E. and Stoica, Ion and Xing, Eric P.},
    month = {March},
    year = {2023}
}

@article{llama3,
  title={The llama 3 herd of models},
  author={Dubey, Abhimanyu and Jauhri, Abhinav and Pandey, Abhinav and Kadian, Abhishek and Al-Dahle, Ahmad and Letman, Aiesha and Mathur, Akhil and Schelten, Alan and Yang, Amy and Fan, Angela and others},
  journal={arXiv preprint arXiv:2407.21783},
  year={2024}
}

@inproceedings{sam,
  title={Segment anything},
  author={Kirillov, Alexander and Mintun, Eric and Ravi, Nikhila and Mao, Hanzi and Rolland, Chloe and Gustafson, Laura and Xiao, Tete and Whitehead, Spencer and Berg, Alexander C and Lo, Wan-Yen and others},
  booktitle={Proceedings of the IEEE/CVF International Conference on Computer Vision},
  pages={4015--4026},
  year={2023}
}

@misc{coyo700m,
  title         = {COYO-700M: Image-Text Pair Dataset},
  author        = {Byeon, Minwoo and Park, Beomhee and Kim, Haecheon and Lee, Sungjun and Baek, Woonhyuk and Kim, Saehoon},
  year          = {2022},
  howpublished  = {\url{https://github.com/kakaobrain/coyo-dataset}},
}

@article{wukong100m,
  title={Wukong: A 100 million large-scale chinese cross-modal pre-training benchmark},
  author={Gu, Jiaxi and Meng, Xiaojun and Lu, Guansong and Hou, Lu and Minzhe, Niu and Liang, Xiaodan and Yao, Lewei and Huang, Runhui and Zhang, Wei and Jiang, Xin and others},
  journal={Advances in Neural Information Processing Systems},
  volume={35},
  pages={26418--26431},
  year={2022}
}

@article{wikipaintings,
  title={Recognizing image style},
  author={Karayev, Sergey and Trentacoste, Matthew and Han, Helen and Agarwala, Aseem and Darrell, Trevor and Hertzmann, Aaron and Winnemoeller, Holger},
  journal={arXiv preprint arXiv:1311.3715},
  year={2013}
}

@article{vsibench,
  title={Thinking in space: How multimodal large language models see, remember, and recall spaces},
  author={Yang, Jihan and Yang, Shusheng and Gupta, Anjali W and Han, Rilyn and Fei-Fei, Li and Xie, Saining},
  journal={arXiv preprint arXiv:2412.14171},
  year={2024}
}

@article{ppocrv3,
  title={PP-OCRv3: More attempts for the improvement of ultra lightweight OCR system},
  author={Li, Chenxia and Liu, Weiwei and Guo, Ruoyu and Yin, Xiaoting and Jiang, Kaitao and Du, Yongkun and Du, Yuning and Zhu, Lingfeng and Lai, Baohua and Hu, Xiaoguang and others},
  journal={arXiv preprint arXiv:2206.03001},
  year={2022}
}

@inproceedings{tang2023visual,
  title={Visual recognition by request},
  author={Tang, Chufeng and Xie, Lingxi and Zhang, Xiaopeng and Hu, Xiaolin and Tian, Qi},
  booktitle={Proceedings of the IEEE/CVF Conference on Computer Vision and Pattern Recognition},
  pages={15265--15274},
  year={2023}
}

@article{wang2023all,
  title={The all-seeing project: Towards panoptic visual recognition and understanding of the open world},
  author={Wang, Weiyun and Shi, Min and Li, Qingyun and Wang, Wenhai and Huang, Zhenhang and Xing, Linjie and Chen, Zhe and Li, Hao and Zhu, Xizhou and Cao, Zhiguo and others},
  journal={arXiv preprint arXiv:2308.01907},
  year={2023}
}

@inproceedings{wang2024all,
  title={The all-seeing project v2: Towards general relation comprehension of the open world},
  author={Wang, Weiyun and Ren, Yiming and Luo, Haowen and Li, Tiantong and Yan, Chenxiang and Chen, Zhe and Wang, Wenhai and Li, Qingyun and Lu, Lewei and Zhu, Xizhou and others},
  booktitle={European Conference on Computer Vision},
  pages={471--490},
  year={2024},
  organization={Springer}
}

@inproceedings{peebles2023scalable,
  title={Scalable diffusion models with transformers},
  author={Peebles, William and Xie, Saining},
  booktitle={Proceedings of the IEEE/CVF international conference on computer vision},
  pages={4195--4205},
  year={2023}
}

@article{wang2023modelscope,
  title={Modelscope text-to-video technical report},
  author={Wang, Jiuniu and Yuan, Hangjie and Chen, Dayou and Zhang, Yingya and Wang, Xiang and Zhang, Shiwei},
  journal={arXiv preprint arXiv:2308.06571},
  year={2023}
}

@inproceedings{mmbench,
  title={Mmbench: Is your multi-modal model an all-around player?},
  author={Liu, Yuan and Duan, Haodong and Zhang, Yuanhan and Li, Bo and Zhang, Songyang and Zhao, Wangbo and Yuan, Yike and Wang, Jiaqi and He, Conghui and Liu, Ziwei and others},
  booktitle={European conference on computer vision},
  pages={216--233},
  year={2024},
  organization={Springer}
}

@article{seedbench,
  title={Seed-bench: Benchmarking multimodal llms with generative comprehension},
  author={Li, Bohao and Wang, Rui and Wang, Guangzhi and Ge, Yuying and Ge, Yixiao and Shan, Ying},
  journal={arXiv preprint arXiv:2307.16125},
  year={2023}
}

@inproceedings{mvbench,
  title={Mvbench: A comprehensive multi-modal video understanding benchmark},
  author={Li, Kunchang and Wang, Yali and He, Yinan and Li, Yizhuo and Wang, Yi and Liu, Yi and Wang, Zun and Xu, Jilan and Chen, Guo and Luo, Ping and others},
  booktitle={Proceedings of the IEEE/CVF Conference on Computer Vision and Pattern Recognition},
  pages={22195--22206},
  year={2024}
}

@article{videomme,
  title={Video-mme: The first-ever comprehensive evaluation benchmark of multi-modal llms in video analysis},
  author={Fu, Chaoyou and Dai, Yuhan and Luo, Yongdong and Li, Lei and Ren, Shuhuai and Zhang, Renrui and Wang, Zihan and Zhou, Chenyu and Shen, Yunhang and Zhang, Mengdan and others},
  journal={arXiv preprint arXiv:2405.21075},
  year={2024}
}

@article{geneval,
  title={Geneval: An object-focused framework for evaluating text-to-image alignment},
  author={Ghosh, Dhruba and Hajishirzi, Hannaneh and Schmidt, Ludwig},
  journal={Advances in Neural Information Processing Systems},
  volume={36},
  pages={52132--52152},
  year={2023}
}

@inproceedings{vbench,
  title={Vbench: Comprehensive benchmark suite for video generative models},
  author={Huang, Ziqi and He, Yinan and Yu, Jiashuo and Zhang, Fan and Si, Chenyang and Jiang, Yuming and Zhang, Yuanhan and Wu, Tianxing and Jin, Qingyang and Chanpaisit, Nattapol and others},
  booktitle={Proceedings of the IEEE/CVF Conference on Computer Vision and Pattern Recognition},
  pages={21807--21818},
  year={2024}
}

@article{t2vcompbench,
  title={T2v-compbench: A comprehensive benchmark for compositional text-to-video generation},
  author={Sun, Kaiyue and Huang, Kaiyi and Liu, Xian and Wu, Yue and Xu, Zihan and Li, Zhenguo and Liu, Xihui},
  journal={arXiv preprint arXiv:2407.14505},
  year={2024}
}

@article{arnab2025temporal,
  title={Temporal Chain of Thought: Long-Video Understanding by Thinking in Frames},
  author={Arnab, Anurag and Iscen, Ahmet and Caron, Mathilde and Fathi, Alireza and Schmid, Cordelia},
  journal={arXiv preprint arXiv:2507.02001},
  year={2025}
}

@inproceedings{wei2024dreamvideo,
  title={Dreamvideo: Composing your dream videos with customized subject and motion},
  author={Wei, Yujie and Zhang, Shiwei and Qing, Zhiwu and Yuan, Hangjie and Liu, Zhiheng and Liu, Yu and Zhang, Yingya and Zhou, Jingren and Shan, Hongming},
  booktitle={Proceedings of the IEEE/CVF Conference on Computer Vision and Pattern Recognition},
  pages={6537--6549},
  year={2024}
}

@article{wei2025dreamrelation,
  title={Dreamrelation: Relation-centric video customization},
  author={Wei, Yujie and Zhang, Shiwei and Yuan, Hangjie and Gong, Biao and Tang, Longxiang and Wang, Xiang and Qiu, Haonan and Li, Hengjia and Tan, Shuai and Zhang, Yingya and others},
  journal={arXiv preprint arXiv:2503.07602},
  year={2025}
}

@article{chang2024survey,
  title={A survey on evaluation of large language models},
  author={Chang, Yupeng and Wang, Xu and Wang, Jindong and Wu, Yuan and Yang, Linyi and Zhu, Kaijie and Chen, Hao and Yi, Xiaoyuan and Wang, Cunxiang and Wang, Yidong and others},
  journal={ACM transactions on intelligent systems and technology},
  volume={15},
  number={3},
  pages={1--45},
  year={2024},
  publisher={ACM New York, NY}
}

@inproceedings{mmmu,
  title={Mmmu: A massive multi-discipline multimodal understanding and reasoning benchmark for expert agi},
  author={Yue, Xiang and Ni, Yuansheng and Zhang, Kai and Zheng, Tianyu and Liu, Ruoqi and Zhang, Ge and Stevens, Samuel and Jiang, Dongfu and Ren, Weiming and Sun, Yuxuan and others},
  booktitle={Proceedings of the IEEE/CVF Conference on Computer Vision and Pattern Recognition},
  pages={9556--9567},
  year={2024}
}

@inproceedings{bao2024cores,
  title={Cores: Orchestrating the dance of reasoning and segmentation},
  author={Bao, Xiaoyi and Sun, Siyang and Ma, Shuailei and Zheng, Kecheng and Guo, Yuxin and Zhao, Guosheng and Zheng, Yun and Wang, Xingang},
  booktitle={European Conference on Computer Vision},
  pages={187--204},
  year={2024},
  organization={Springer}
}

@article{bao2025dynimg,
  title={DynImg: Key Frames with Visual Prompts are Good Representation for Multi-Modal Video Understanding},
  author={Bao, Xiaoyi and Xie, Chenwei and Tang, Hao and Weng, Tingyu and Wang, Xiaofeng and Zheng, Yun and Wang, Xingang},
  journal={arXiv preprint arXiv:2507.15569},
  year={2025}
}

@inproceedings{liu2025hybrid,
  title={Hybrid-level instruction injection for video token compression in multi-modal large language models},
  author={Liu, Zhihang and Xie, Chen-Wei and Li, Pandeng and Zhao, Liming and Tang, Longxiang and Zheng, Yun and Liu, Chuanbin and Xie, Hongtao},
  booktitle={Proceedings of the Computer Vision and Pattern Recognition Conference},
  pages={8568--8578},
  year={2025}
}

@article{deepeyes,
  title={DeepEyes: Incentivizing" Thinking with Images" via Reinforcement Learning},
  author={Zheng, Ziwei and Yang, Michael and Hong, Jack and Zhao, Chenxiao and Xu, Guohai and Yang, Le and Shen, Chao and Yu, Xing},
  journal={arXiv preprint arXiv:2505.14362},
  year={2025}
}

@article{jiang2025vlm,
  title={VLM-R3: Region Recognition, Reasoning, and Refinement for Enhanced Multimodal Chain-of-Thought},
  author={Jiang, Chaoya and Heng, Yongrui and Ye, Wei and Yang, Han and Xu, Haiyang and Yan, Ming and Zhang, Ji and Huang, Fei and Zhang, Shikun},
  journal={arXiv preprint arXiv:2505.16192},
  year={2025}
}

@inproceedings{jiang2023efficient,
  title={Efficient decision-based black-box patch attacks on video recognition},
  author={Jiang, Kaixun and Chen, Zhaoyu and Huang, Hao and Wang, Jiafeng and Yang, Dingkang and Li, Bo and Wang, Yan and Zhang, Wenqiang},
  booktitle={Proceedings of the IEEE/CVF International Conference on Computer Vision},
  pages={4379--4389},
  year={2023}
}

@inproceedings{liu2024towards,
  title={Towards balanced alignment: Modal-enhanced semantic modeling for video moment retrieval},
  author={Liu, Zhihang and Li, Jun and Xie, Hongtao and Li, Pandeng and Ge, Jiannan and Liu, Sun-Ao and Jin, Guoqing},
  booktitle={Proceedings of the AAAI conference on artificial intelligence},
  volume={38},
  pages={3855--3863},
  year={2024}
}

@article{jiang2025videopure,
  title={VideoPure: Diffusion-based Adversarial Purification for Video Recognition},
  author={Jiang, Kaixun and Chen, Zhaoyu and Fu, Jiyuan and Hong, Lingyi and Li, Jinglun and Zhang, Wenqiang},
  journal={IEEE Transactions on Circuits and Systems for Video Technology},
  year={2025},
  publisher={IEEE}
}

@article{allen1955machine,
  title={Machine literature searching VIII. Operational criteria for designing information retrieval systems},
  author={Allen, Kent and Berry, Madeline M and Luehrs Jr, Fred U and Perry, James W},
  journal={American Documentation (pre-1986)},
  volume={6},
  number={2},
  pages={93},
  year={1955},
  publisher={Wiley Periodicals Inc.}
}

@inproceedings{chinchor1998appendix,
  title={Appendix b: Muc-7 test scores introduction},
  author={Chinchor, Nancy},
  booktitle={Seventh Message Understanding Conference (MUC-7): Proceedings of a Conference Held in Fairfax, Virginia, April 29-May 1, 1998},
  year={1998}
}
